\documentclass[a4paper,fleqn]{cas-dc}

\usepackage{float}
\usepackage[numbers]{natbib}

\usepackage{tabularx}

\def\tsc#1{\csdef{#1}{\textsc{\lowercase{#1}}\xspace}}
\tsc{WGM}
\tsc{QE}
\tsc{EP}
\tsc{PMS}
\tsc{BEC}
\tsc{DE}

\DeclareUnicodeCharacter{2212}{-}

\begin{document}
\let\WriteBookmarks\relax
\def\floatpagepagefraction{1}
\def\textpagefraction{.001}

\shorttitle{Small and Dim Target Detection in IR Imagery: A Review}

\shortauthors{Nikhil et~al.}

\title [mode = title]{Small and Dim Target Detection in IR Imagery: A Review} 




%
\author[1,2]{Nikhil Kumar}[]



\ead{nikhil_k1@cs.iitr.ac.in}



\author[1]{Pravendra Singh}[orcid=0000-0003-1001-2219]
\cormark[1]
\ead{pravendra.singh@cs.iitr.ac.in}


\cortext[cor1]{Corresponding author.}

\affiliation[1]{
    addressline={Department of Computer Science and Engineering, Indian Institute of Technology Roorkee}, 
    city={Roorkee},
    citysep={}, 
    postcode={247667}, 
     state={Uttarakhand},
    country={India}}

\affiliation[2]{
    addressline={Instruments Research and Development Establishment, Defence Research and Development Organization}, 
    city={Dehradun},
    citysep={}, 
    postcode={248008}, 
     state={Uttarakhand},
    country={India}}




\begin{abstract}
While there has been significant progress in object detection using conventional image processing and machine learning algorithms, exploring small and dim target detection in the IR domain is a relatively new area of study. The majority of small and dim target detection methods are derived from conventional object detection algorithms, albeit with some alterations. The task of detecting small and dim targets in IR imagery is complex. This is because these targets often need distinct features, the background is cluttered with unclear details, and the IR signatures of the scene can change over time due to fluctuations in thermodynamics. The primary objective of this review is to highlight the progress made in this field. This is the first review in the field of small and dim target detection in infrared imagery, encompassing various methodologies ranging from conventional image processing to cutting-edge deep learning-based approaches. The authors have also introduced a taxonomy of such approaches. There are two main types of approaches: methodologies using several frames for detection, and single-frame-based detection techniques. Single frame-based detection techniques encompass a diverse range of methods, spanning from traditional image processing-based approaches to more advanced deep learning methodologies. Our findings indicate that deep learning approaches perform better than traditional image processing-based approaches. In addition, a comprehensive compilation of various available datasets has also been provided. Furthermore, this review identifies the gaps and limitations in existing techniques, paving the way for future research and development in this area. 


\end{abstract}


\maketitle

\section{Introduction}
Infrared (IR) imagery has emerged as a pivotal technology in various fields, including surveillance, reconnaissance, and target detection. The ability to capture thermal radiation emitted by objects enables the acquisition of valuable information, especially in scenarios where traditional optical sensors may fall short \cite{kou2023detection}. One of the critical challenges within the realm of IR imagery is the detection of small and dim targets \cite{liu2023infrared, zhang2023dim2clear, cao2023infrared, lu2023infrared}, a task that demands advanced techniques and sophisticated algorithms.

The identification of small and dim targets in IR imagery holds paramount significance in applications such as military surveillance, search and rescue operations, and environmental monitoring \cite{kou2022infrared, zhang2024global}. These targets, often characterized by their low thermal contrast against the background, pose a formidable challenge for traditional image processing methods. The complexity arises from factors such as noise, clutter, and varying environmental conditions, all of which can obscure the detection of these subtle thermal signatures.

As technology advances, the demand for robust and efficient small and dim target detection algorithms becomes increasingly pressing. Addressing this challenge requires a multidisciplinary approach, combining expertise in signal processing, machine learning, deep learning, and computer vision. Researchers are driven to develop innovative methodologies \cite{zhao2022single, rawat2020review} that enhance the sensitivity and accuracy of IR imagery systems, enabling the reliable detection of elusive targets that may otherwise escape notice.

The main objective of this review is to offer an extensive evaluation of the advancements in the realm of detecting small and dim targets within IR domain. This assessment covers both conventional techniques rooted in image processing and more sophisticated methods founded on deep learning. By scrutinizing various approaches, our aim is to shed light on the advancements made in detecting diminutive and faint objects, while also gaining a comprehensive understanding of the strengths and limitations inherent in these methods. Additionally, this paper delivers a thorough examination of currently available datasets designed for detecting small and dim targets in IR images.

\subsection{IR imaging}
Most objects with temperatures above absolute zero emit a notable amount of IR radiation, as outlined by \cite{hudson1969infrared}. Such sources are all around us. In the creation of IR images, a substantial portion of the signals consist of radiated emissions. One notable advantage of these emissions is their persistent presence and their ability to withstand deterioration caused by adverse weather conditions. These characteristics make the IR spectrum a practical wavelength for imaging, especially in applications related to defense.
The origins of IR radiation can be traced back to a scientific experiment conducted by Frederick William Herschel over two centuries ago. Herschel used prisms and basic temperature sensors to study the distribution of wavelengths across the electromagnetic spectrum. It is now widely accepted that objects emit radiation across a wide spectrum of wavelengths, following established principles of physics. Within the electromagnetic spectrum, the infrared (IR) region spans wavelengths from 700 nanometers to 1 millimeter, with the lower end coinciding with the red edge of the visible spectrum. Empirical evidence shows that a significant portion of the IR spectrum is unsuitable for conventional applications due to the absorption of IR radiation by atmospheric water or carbon dioxide \cite{ holst2000common,  lloyd2013thermal}. This IR range is conventionally divided into five spectral sub-bands. The near-infrared (NIR) spectrum covers wavelengths from 0.7 µm to 0.9 µm, while the shortwave IR (SWIR) ranges from 0.9 µm to 2.5 µm. The mid-wave infrared (MWIR) spectrum includes wavelengths between 3 µm and 5 µm, and the long-wave infrared (LWIR) spectrum comprises wavelengths from 8 µm to 12 µm. Lastly, the far-infrared (FIR) spectrum extends to wavelengths of up to 1000 micrometers.

\subsection{Significance of Small and Dim targets}

As defined by the International Society for Optics and Photonics (SPIE) and elaborated in Zhang's work \cite{zhang2003algorithms}, a small target is one that occupies less than 0.12 percent of the total pixels in an image. In the context of 256x256 images, targets smaller than 80 pixels are categorized as small.

In military operations, many aerial targets exhibit distinct signatures within the medium-wave infrared (MWIR) band, making it a critical band of operation. Consequently, various electro-optical systems like IR Search and Track (IRST) and Missile Approach Warning Systems (MAWS), as referenced in the works of \cite{de1995irst} and \cite{holm2010missile}, are designed to operate within this segment of the electromagnetic spectrum. In addition to this, the detection, identification, and tracking of small and dim objects play a crucial role in the field of infrared guidance and unmanned aerial vehicles (UAVs) \cite{zhang2024global}.
In numerous situations, these targets need to be detected at considerable distances, which can extend to several hundred kilometers. As a result, an infrared sensor \cite{kruse2001uncooled} will only be able to see distant targets with small angular sizes and few pixel-based target signatures. In airborne scenarios, military targets, despite emitting strong infrared radiation, often appear faint in the image plane due to significant transmission losses, including absorption and scattering \cite{hudson1969infrared}, particularly over long distances.
Identifying and tracking such small targets holds great importance in defense applications. Infrared imagery frequently suffers from notable noise and background clutter, causing targets to become obscured, reducing contrast and diminishing the signal-to-clutter ratio (SCR). Consequently, aerial targets of military significance are often observed as small and faint targets.

\subsection{Why Specialised Algorithms for Detecting Small and Dim Targets in IR Imagery}

Small IR targets often have small dimensions, low intensity, amorphous structures, and lack texture, making them vulnerable to blending into complex background clutter. The direct application of popular generic deep learning-based object detection algorithms like the RCNN series \cite{girshick2014rich}, YOLO series \cite{redmon2016you}, and SSD \cite{liu2016ssd} for detecting small and dim point targets is not suitable because pooling layers in these networks may result in the loss of such targets in deeper layers. Researchers have focused on developing deep networks customized for detecting small IR targets by leveraging domain-specific knowledge.
In contrast to techniques for small target detection in RGB images, which primarily address the issue of small target size \cite{uzair2020bio, zhou2020hierarchical} and employ strategies like context information learning \cite{torralba2001statistical}, data augmentation \cite{zoph2020learning} and multi-scale learning \cite{singh2018analysis}   to enhance detection robustness and generalization, applying these techniques directly to the detection of small and dim targets in IR imagery, as indicated by in \cite{wang2019miss}, leads to a significant drop in performance.
Convolutional neural networks (CNNs) typically include max-pooling layers, which have the potential to suppress or eliminate small and dim IR targets, as observed in Liangkui's work \cite{liangkui2018using}. Hence, there's a need for specialized neural network architectures to effectively address these specific challenges.

\subsection{Challenges}
Long-range surveillance systems employed in defense applications frequently rely on MWIR imaging systems, as discussed in Singh's book \cite{singh2009thermal}. In such scenarios, the signatures of objects typically occupy only a small number of pixels, resulting in limited spatial features.
The infrared signatures are susceptible to temporal fluctuations due to the dynamic nature of scene thermodynamics and changes in the visual aspect angle of the observed object in relation to the imaging system's position.
Infrared imagery often suffers from high levels of noise and background clutter, which can obscure targets, reduce contrast, and result in a diminished Signal-to-Clutter Ratio (SCR). IR small-dim targets exhibit a higher resemblance to the background with a low SCR compared to targets in RGB images, making it more challenging to distinguish small-dim targets from their surroundings.

\subsection{Motivation and Contribution}
A significant proportion of long-distance objects observed by aerial targets tend to be small targets. This characteristic underscores the importance of studying the detection of such targets, particularly within defense applications. Though recent research efforts have concentrated on this field, it seems that the scope is quite broad within the IR domain. These methods encompass a range from conventional image processing approaches to cutting-edge deep learning methods.

\par
\begin{figure*}{}
	\centerline{\includegraphics[scale=0.8]{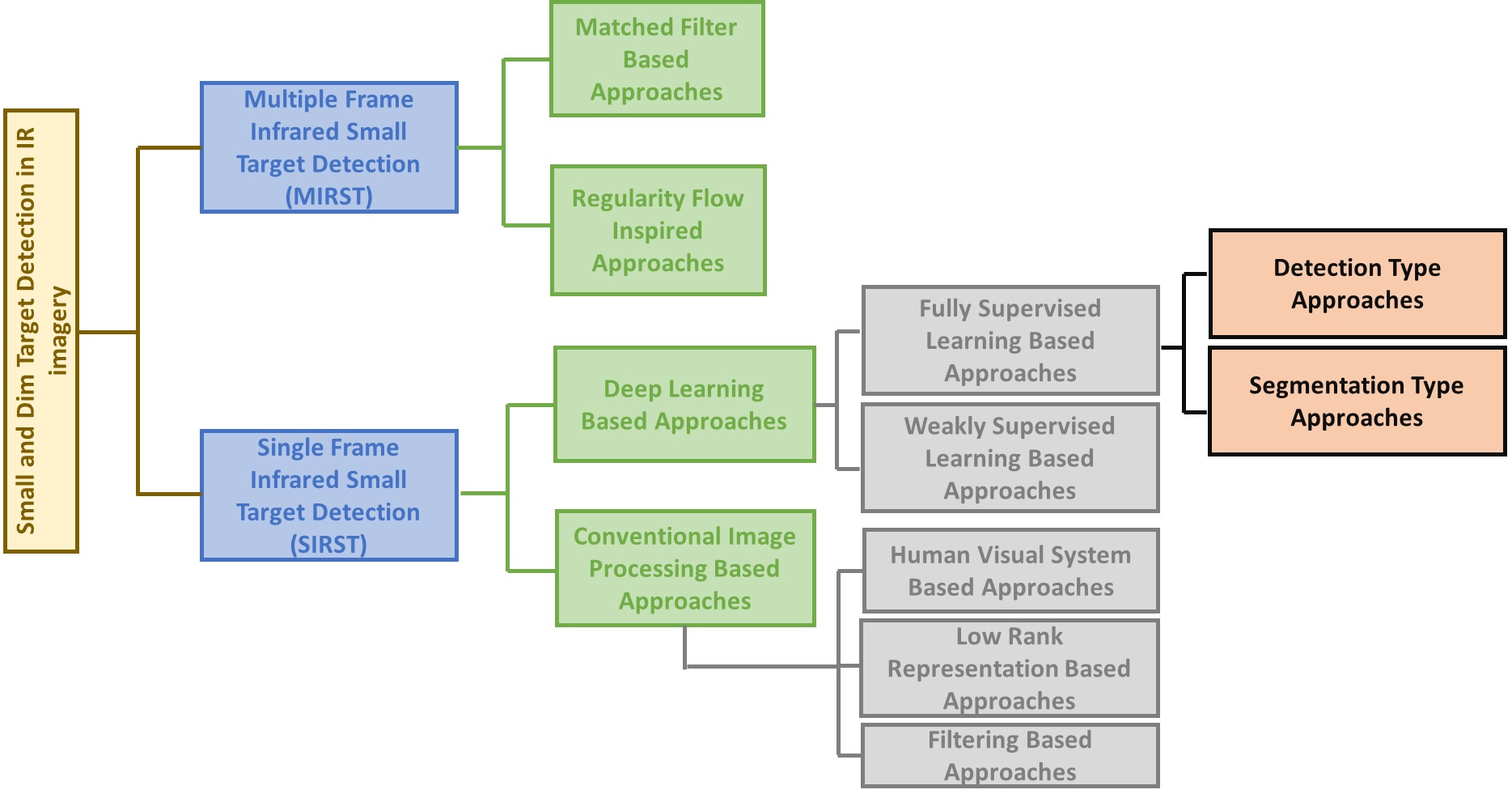}}
	\caption{Classification of approaches for small and dim target detection in infrared imagery.}
	\label{class}
\end{figure*}

 Recently, only a limited number of reviews in this field have been published. Zhao et al.'s study \cite{zhao2022single} specifically focuses on single-frame infrared small-target detection approaches. Rawat et al. \cite{rawat2020review} exclusively concentrate on traditional image processing approaches, while Kou et al. \cite{kou2023infrared} solely address machine learning-based approaches. The present work represents the first comprehensive survey in this field, examining various technologies for detecting small and dim targets in infrared imagery. Our survey covers a spectrum of approaches, from conventional image processing to cutting-edge deep learning methods. Additionally, our work incorporates up-to-date approaches in this area. The classification illustrated in Figure \ref{class} outlines the various approaches employed for the detection of small and dim targets in IR imagery. Small and dim target detection algorithms can be primarily categorized into two groups based on their implementation: Multiple frame InfraRed Small Target (MIRST) and Single frame InfraRed Small Target (SIRST). In terms of computational complexity, SIRST techniques are considered to be more advanced alternatives. SIRST methods are further divided into conventional image processing-based and deep learning-based approaches. Our survey paper also offers further classification of  SIRST and MIRST methodologies. A comprehensive compilation of most of the datasets pertaining to this area has also been presented.

The remainder of this review is organized as follows. Sections \ref{sec:two} and \ref{sec:three} provide a detailed taxonomy of these approaches. Section \ref{sec:four} offers a comprehensive overview of the datasets available for the specific purpose of detecting small and faint targets in IR imagery. Performance metrics relevant to the problem settings are discussed in Section \ref{sec:five}, while a detailed discussion of the performance of these techniques, including potential future directions, is presented in Section \ref{sec:six}. The conclusion is presented in Section \ref{sec:seven}.

\section {Multiple frame InfraRed Small Target (MIRST) Detection} \label{sec:two}

In these approaches, multiple frames are simultaneously employed for the purpose of detecting targets. The majority of MIRST detection algorithms described in the literature utilize conventional image processing techniques. The following are the representative methods falling under this category.

\subsection{Matched Filter Based Approaches}

\par Reed et al. \cite{reed1988optical} introduced the utilization of three-dimensional matched filtering as a robust processing technique for detecting weak,  targets in motion within a noisy background. This method involves the manipulation of complete sequences of optical frames that encompass mobile targets. This process necessitates precise matching with the target's signature and velocity vector, allowing it to simultaneously detect all matched targets. However, a primary challenge associated with 3-D matched filtering is the need to match the filter to a specific velocity profile. This means that the filter must be customized for a predefined target moving at a particular velocity and direction. To address this limitation to some extent, a filter bank can be implemented, tailored to encompass the targeted speed and direction uncertainties.

\par Porat and Friedland \cite{porat1990frequency} tackled this issue by employing a bank of 3-dimensional Directional Derivative Filters (3DDFs). They examined each possible target direction separately and devised a rule to determine the presence or absence of targets in each direction. Their findings indicate that the signal-to-noise ratio (SNR) increases linearly as the integration time extends. This increase surpasses what is typically achieved through the application of 2-dimensional matched filtering to a single frame. The authors presented their results graphically and the accuracy of the plotted SNR can be readily verified through direct computation, using fundamental calculations of signal and noise power. This aligns with the \textit{theoretical} expectations.

 \par Li et al. \cite{li2005three} introduced a 3D Double Directional Filter (3DDDF) based approach for detecting and tracking small moving targets within complex and cluttered backgrounds in sequences of IR images. This algorithm employs a double-directional filtering technique in three dimensions, enhancing the target's energy accumulation beyond that of the 3DDF method. Before applying the filter, they employ a pre-whitening technique known as a Three-Dimensional Spatial-Temporal Adaptive Prediction Filter (TDSTAPF) to mitigate the effects of cluttered backgrounds. Comprehensive experiments presented by Li et al. \cite{li2005three} have demonstrated that their algorithms are capable of detecting weak dim point targets amidst complex backgrounds cluttered with clouds in real IR image sequences. The authors conducted a performance assessment of the 3DDDF technique in practical settings using infrared image sequences. They also compared the performance of the 3DDDF algorithm with the 3DDF algorithm, using actual IR image data. The experimental results suggest that the 3DDDF algorithms outperform the 3DDF algorithms.

\par One of the primary challenges associated with methodologies that rely on information from multiple frames is their limited effectiveness when applied with a moving camera. In such cases, an additional pre-processing step, such as image registration, becomes necessary, rendering these approaches more computationally expensive.

\subsection{Regularity Flow-based Approaches}

In this category of approaches, spatio-temporal regularity is considered a prominent feature and its pattern is analyzed in the temporal domain. The following is a representative algorithm within this category.

\par Nikhil et al. \cite{kumar2016regularity} introduced a Hough transform \cite{hough1962method} based approach for detecting small and dim targets. This method is based on the formation of a video data cuboid. Instead of detecting targets in the $X-Y$ plane, it explores the trajectories of targets in $X-T$ slices. One of the primary assumptions of this method is a stationary camera. Due to the greater coverage of pixels in the $X-T$ plane by the trajectories of small targets compared to the number of target pixels in the $X-Y$ plane, there is an inherent increase in the signal-to-noise ratio (SNR). The authors compiled a dataset that includes multiple mobile vehicles captured from a distance of approximately 5 kilometers using a MWIR imager. As a result of the significant distance, the majority of captured objects often appear as small targets. Their findings demonstrate exceptional performance, particularly in effectively managing occlusion and rejecting clutter. It's worth noting that all video sequences analyzed in their study exclusively capture sequences of ground-based targets and do not include any aerial targets.

\section {Single-frame InfraRed Small Target (SIRST) Detection} \label{sec:three}

Within this specific domain, researchers have employed a range of innovative methodologies. These methods can be further categorized into two subcategories: conventional image processing-based approaches and deep learning-based approaches.

\subsection{Conventional Image Processing Based Approaches }

Algorithms falling under this category tend to have low computational complexity, lack generalizability, and struggle to suppress non-uniform backgrounds effectively. They may also struggle with complex backgrounds, resulting in low detection rates and inadequate stability. These approaches can be further classified into three main categories: Low-rank Representation-based approaches, Human Visual System (HVS) based approaches, and Filtering-based approaches.

\subsubsection{Low-rank Representation Based Approaches}
These approaches leverage the mathematical properties of matrices and employ an alternative low-dimensional representation for the purpose of detecting small and dim targets in IR imagery.

\par The authors \cite{gao2013infrared} introduced a technique known as the IR Patch-Image model (IPI), which utilizes a patch-based image model to enhance the accuracy of small target detection. This method involves breaking down the input IR image into numerous overlapping patches and creating a model that captures the relationship between these patches and the overall image. By employing a non-local self-similarity constraint, the model can capture the inherent similarity between patches, which aids in the detection process.
In the IPI model, the IR image can be mathematically expressed as shown in Equation \ref{eq:ipi1}:
\begin{equation}
f_D(x, y) = f_T(x, y) + f_B(x, y) + f_N(x, y)
\label{eq:ipi1}
\end{equation}
Here, the variables $f_D$, $f_T$, $f_B$, $f_N$, and $(x, y)$ represent the original IR image, the target image, the background image, the random noise image, and the pixel location, respectively. In IPI model, corresponding patch-images $D$, $B$, $T$ and $N$ can be constructed and expressed as shown in Equation \ref{eq:ipi2}:
\begin{equation}
D = B + T + N
\label{eq:ipi2}
\end{equation}
The patch image $D$ is generated using the original IR image $f_D$, which is derived from an image sequence. The Accelerated Proximal Gradient algorithm is utilized to simultaneously estimate the low-rank background patch-image $B$ and the sparse target patch-image $T$ within the patch-image $D$. Subsequently, the process of reconstructing the background image $f_B$ and the target image $f_T$ is carried out by using the patch images $B$ and $T$, respectively.
An algorithm employing a straightforward segmentation technique is applied to dynamically partition the target image $f_T$, responding to the presence of minor errors characterized by low magnitudes. Finally, through post-processing techniques, the segmentation results are optimized to achieve the final detection outcome.
The test image synthesis involves using actual IR background images along with various targets. The background images are selected from multiple real image sequences with varying levels of clutter. The different targets are created by resizing four real targets using the bi-cubic interpolation method. The dataset is divided into ten groups, each with different target sizes ranging from 5 to 37 pixels and varying numbers of targets. It has been observed that the probability of detection can vary from 0.5 to 0.9, with an average value of 0.82.

\par The authors \cite{dai2017reweighted} introduced a technique known as the Reweighted IR Patch-Tensor model (RIPT) to effectively leverage both local and non-local priors simultaneously. Initially, they used the IR Patch-Tensor (IPT) model to accurately represent the image while preserving its spatial correlations. By incorporating the sparse prior of the target and the non-local self-correlation prior to the background, the authors formulated the task of separating the target from the background as a robust low-rank tensor recovery problem.
To integrate the local structure prior to the IPT model, the authors created a weight for each element based on the structure tensor. This weight is designed to reduce the influence of remaining edges and maintain the desired target dimensionality. To enhance computational efficiency, a re-weighted scheme was implemented to increase the sparsity of the patch tensor of the target. Due to the unique nature of IR small target detection, an additional stopping criterion was implemented to prevent excessive computational processing.
The authors computed the local structure feature map of an IR image and generated the original patch-tensor and the local structure weight patch-tensor using the IR image and the local structure map. They decomposed the patch-tensor into two separate tensors: the background patch-tensor and the target patch-tensor. The background image and target image were reconstructed using these tensors, and the target was segmented using a methodology similar to the one described in  IPI model  \cite{gao2013infrared}.
The performance of this algorithm was evaluated using the NUAA-SIRST \cite{dai2021asymmetric} and IRSTD \cite{zhang2022isnet} datasets.

\begin{table*}[t]
\caption{Performance comparison of low-rank representation-based algorithms.}
\label{tab:LR}
\begin{tabular}{@{}lllllll@{}}
\toprule
\multicolumn{1}{c}{\textbf{Reference}} & \multicolumn{1}{c}{\textbf{Method}} & \multicolumn{1}{c}{\textbf{Type}}              & \multicolumn{1}{c}{\textbf{Dataset}} & \multicolumn{3}{c}{\textbf{Result}}          \\ \midrule
\multirow{6}{*}{\cite{gao2013infrared}}                      & \multirow{6}{*}{IPI}                & \multirow{6}{*}{Low Rank Representation Based} & \multirow{3}{*}{NUDT-SIRST}          & Pixel Level                   & IOU & 17.76 \\  \\
                                       &                                     &                                                &                                      & \multirow{2}{*}{Object Level} & Pd  & 74.79  \\
                                       &                                     &                                                &                                      &                               & Fa  & 41.23  \\ \\
                                       &                                     &                                                & \multirow{3}{*}{NUAA-SIRST}          & Pixel Level                   & IOU & 25.67   \\ \\
                                       &                                     &                                                &                                      & \multirow{2}{*}{Object Level} & Pd  & 85.55  \\
                                       &                                     &                                                &                                      &                               & Fa  & 11.47   \\ \\ \midrule
\multirow{6}{*}{\cite{dai2017reweighted}}                      & \multirow{6}{*}{RIPT}               & \multirow{6}{*}{Low Rank Representation Based} & \multirow{3}{*}{NUDT-SIRST}          & Pixel Level                   & IOU & 29.44 \\ \\
                                       &                                     &                                                &                                      & \multirow{2}{*}{Object Level} & Pd  & 91.85   \\
                                       &                                     &                                                &                                      &                               & Fa  & 344.33  \\ \\
                                       &                                     &                                                & \multirow{3}{*}{NUAA-SIRST}          & Pixel Level                   & IOU & 11.05   \\ \\
                                       &                                     &                                                &                                      & \multirow{2}{*}{Object Level} & Pd  & 79.08  \\
                                       &                                     &                                                &                                      &                               & Fa  & 22.61  \\  \\
                                       \midrule
\end{tabular}

\end{table*}

Table \ref{tab:LR} illustrates the performance comparison of Low-Rank Representation-based algorithms on the NUDT-SIRST and NUAA-SIRST datasets. Methods employing low-rank representation have the capability to adapt to low SCR in IR images. However, they still face the significant challenge of generating a high number of false alarms when applied to images that contain small and irregularly shaped targets within complex backgrounds.

\subsubsection{Human Visual System (HVS) Based Approach}

HVS-based algorithms leverage principles of human visual perception, encompassing low-level image processing and higher-level cognitive processes, to extract relevant information and differentiate point targets from the surrounding background. By incorporating human-like processing capabilities, these algorithms aim to address challenges such as complex backgrounds, limited signal strength relative to noise and variations in the appearance of the target.

\par 
A representative algorithm in this category is the Local Contrast Method (LMS) \cite{chen2013local}. Researchers have concluded that contrast is a fundamental attribute encoded within the streams of the visual system \cite{kim2012scale, vanrullen2003visual}. This holds true throughout the entire target detection process, as mentioned in previous studies. It has been observed that small targets exhibit a distinct pattern of discontinuity compared to their surrounding regions. The target is concentrated within a relatively small area, characterized as a homogeneous compact region \cite{wang1996efficient}. Additionally, the target's background aligns with its neighboring regions consistently. Therefore, it is hypothesized that a local area exhibiting contrast greater than a designated threshold at a specific scale could potentially indicate the presence of the target.

In the algorithm by \cite{chen2013local}, the image is divided into a three-by-three grid of cells. The cell at the center of the grid is labeled as 0, while the remaining cells are labeled as 1, 2, 3, 4, 5, 6, 7, and 8. The values of $L_n$, $m_i$ and $C_n$ are determined using Equation \ref{eq:hvs1}, Equation \ref{eq:hvs2} and  Equation \ref{eq:hvs3}. The value of the center cell is denoted as $C_n$, and a computed threshold $Th$ is applied for segmentation.

\begin{gather}
L_{n}  = max_{j=1,2,...,N_0}  {I_0}^ j\label{eq:hvs1}  \\ 
m_i= \frac{1}{N_u} \sum_1^{N_u} {I_j}^i\label{eq:hvs2} \\ 
C_n = min_i(  \frac{L_{n} ^{2}}{m_i}) \label{eq:hvs3}
\end{gather}

\begin{table*}[t]
\caption{Performance comparison of HVS-based approach.}
\label{tab:my-hvs}
\begin{tabular}{@{}lllcll@{}}
\toprule
\multicolumn{1}{c}{Reference} &
  \multicolumn{1}{c}{Method} &
  \multicolumn{1}{c}{Type} &
  Dataset &
  \multicolumn{2}{c}{Result} \\ \midrule
\multirow{4}{*}{\cite{chen2013local}} &
  \multirow{4}{*}{LCM} &
  \multirow{4}{*}{HVS} &
  \multirow{2}{*}{\begin{tabular}[c]{@{}c@{}}Custom \\ (Gaussian White Noise of Variance 0.00001) \end{tabular}} &
  Pd &
  86.67 \\
 &
   &
   &
   &
  FA &
  0.2883 \\ \\
 &
   &
   &
  
  \multirow{2}{*}{\begin{tabular}[c]{@{}c@{}}Custom\\ (Gaussian White Noise of Variance 0.00005) \end{tabular}} &
  Pd &
  83.33 \\
 &
   &
   &
   &
  FA &
  0.3333 \\ \\ \bottomrule
\end{tabular}
\end{table*}

The performance of the Human Visual System (HVS)-based technique with a customized dataset is shown in Table \ref{tab:my-hvs}. This table presents two scenarios that have been examined, each characterized by Gaussian white noise with standard deviations of 0.00001 and 0.00005, respectively.

In the case of the method utilizing HVS, it has been observed that such a method is not effective in effectively reducing the unwanted elements present in the background. Local contrast-based methodologies demonstrate greater suitability for high-contrast targets as opposed to dim ones.

\subsubsection{Filtering Based Approaches}

In the IR target detection area, these algorithms were some of the earliest solutions designed to address the challenge of identifying small targets in IR images. They operate by analyzing variations in grayscale features and the visual saliency of elements within the image to locate faint and small targets. The process begins with an estimation of the IR background, followed by the implementation of a technique to suppress this background. Subsequently, a decision plane is established to isolate small and faint objects of interest. This procedure can be likened to the operation of a high-pass filter.

\par
\begin{figure}[t]
	\centerline{\includegraphics[scale=0.8]{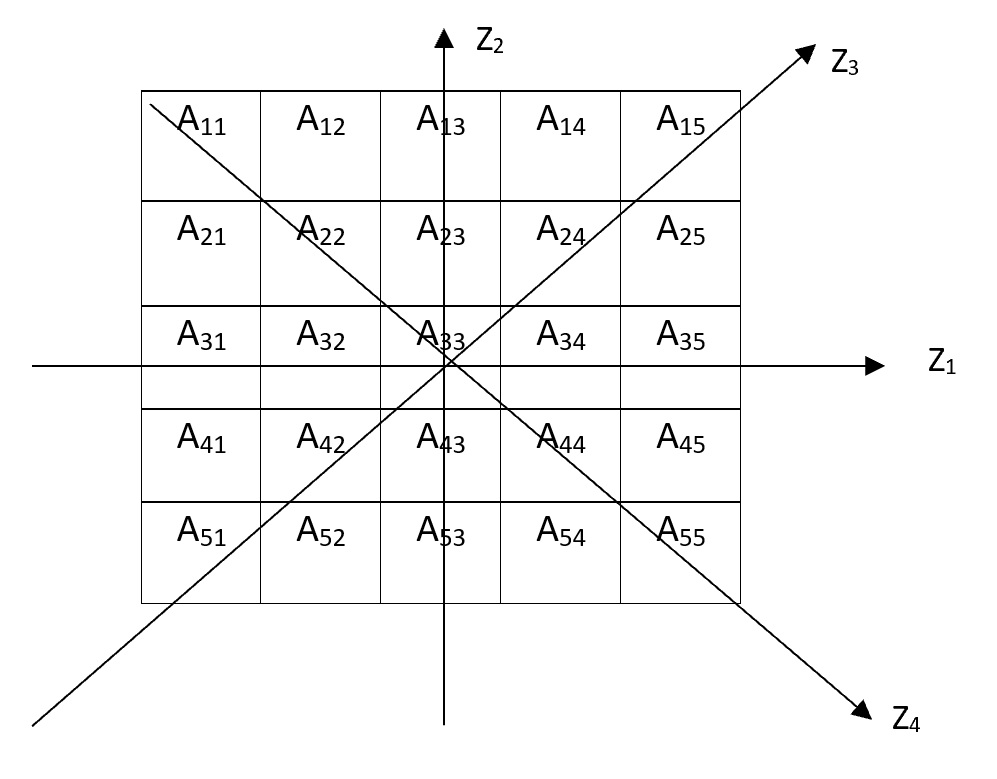}}
	\caption{Max-Mean/Max-Median algorithm.}
	\label{max_mean}
\end{figure}

 The researchers \cite{gu2010kernel} in their work introduced a method referred to as the Max-Mean Filter technique. This method entails the sliding of a window across the image of the scene. Four mean values, denoted as $Z_1, Z_2, Z_3, Z_4$ are computed for the elements within the window in both horizontal and vertical directions, as well as in two diagonal directions. This process is depicted in the accompanying Figure \ref{max_mean} and Equation \ref{eq10}. The central value of the window is then replaced with the highest value among these four computed values.

\begin{gather}
Z_1=mean\{A_{31},A_{32},A_{33},A_{34},A_{35}\}\\ 
Z_2=mean\{A_{13},A_{23},A_{33},A_{43},A_{53}\}\\ 
Z_3=mean\{A_{15},A_{24},A_{33},A_{42},A_{51}\}\\ 
Z_4=mean\{A_{11},A_{22},A_{33},A_{44},A_{55}\} \label{eq10}
\end{gather}
and $max\{Z_1,Z_2,Z_3,Z_4\}$ replaces $A_{33}$.
The choice of window size is a crucial parameter that has a substantial impact on the method's effectiveness. It has been noticed that when we increase the window size at a specific threshold, both the Detection Rate (DR) and the False Alarm Rate (FAR) tend to rise. Hence, when selecting the window size, it's vital to ensure that we attain satisfactory values for both DR and FAR. Achieving the best performance requires finding a suitable balance between these two aspects.

\par The Max-Median algorithm, introduced by the authors \cite{deshpande1999max}, is akin to the Max-Mean method. It involves the movement of a window across an IR image. In this process, the variables $Z_1, Z_2, Z_3, Z_4$ are employed to denote the directions where median values need to be computed for the window elements. These directions encompass both horizontal and vertical orientations, along with two diagonal directions. The central value of the window is subsequently substituted with the maximum value among these four calculated values, as illustrated in Figure \ref{max_mean}.
\begin{gather}
Z_1=median\{A_{31},A_{32},A_{33},A_{34},A_{35}\}\\ 
Z_2=median\{A_{13},A_{23},A_{33},A_{43},A_{53}\}\\ 
Z_3=median\{A_{15},A_{24},A_{33},A_{42},A_{51}\}\\ 
Z_4=median\{A_{11},A_{22},A_{33},A_{44},A_{55}\} \label{eq11}
\end{gather}
Here $A_{33}$ is used to replace by $max\{Z_1,Z_2,Z_3,Z_4\}$.

\par The Top Hat Morphology (THM), as described by Gonzalez in his work \cite{gonzalez2004digital}, is a morphological technique that makes use of the top-hat transform to extract small elements and intricate features from provided images. The top-hat transform has two distinct types: The first type is the white top-hat transform, which entails calculating the discrepancy between the input image and its opening operation using a specific structuring element. The second type is the black top-hat transform, defined as the disparity between the closing operation of the input image and the input image itself. The white top-hat transform is commonly applied in the context of point and small target detection. If we denote a grayscale image as $f$ and a structuring element as $b$, then the white top-hat transform of an image $f$ can be expressed as follows:

\begin{equation}
	T_w (f)=f-f\circ b \label{eq1}
\end{equation}
where $\circ$ denotes the opening operation. The opening operation $f\circ b$ consists of an erosion operation followed by a dilation operation.
\begin{equation}
	f\circ b=\left(  f \ominus b \right)  \oplus b \label{eq2}
\end{equation}
The white top-hat transform generates an image that emphasizes elements in the input image smaller than the specified structuring element. Essentially, it accentuates regions where the structuring element doesn't align and appears brighter than its surroundings. Subsequent to employing the white top-hat transform, it's common to apply a global threshold to divide the detection plane into pixels representing targets and those depicting the background. This thresholding procedure aids in the differentiation of pertinent targets from their adjacent background, simplifying the process of detection and analysis.

\par
\begin{figure}[t]
	\centerline{\includegraphics[scale=0.8]{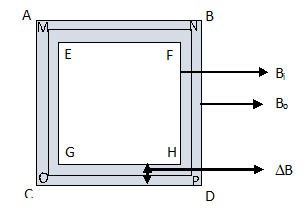}}
	\caption{Structuring element utilized in MTHM.}
	\label{fig1}
\end{figure}

 The authors \cite{bai2008new, bai2010enhancement} introduced a technique known as Modified Top-Hat Morphology (MTHM) to enhance the detection of small targets. Small targets typically exhibit a concentrated and bright region against an IR-cluttered background. The surrounding areas in the image usually have a significant contrast in gray intensity compared to the central region. However, the classical top-hat transformation, which uses identical structuring elements, may not effectively distinguish the target region from the surrounding region. This is due to the limited discriminative power of the structuring elements used.
To address this limitation and optimize the use of gray intensity differences between the target region and its surroundings, the authors propose the use of two structuring elements. This approach can help mitigate the impact of noise and enhance the detection of small targets. Let $B_{oi}$ denote the combination of two structuring elements, namely $B_o, B_i$, as illustrated in Figure \ref{fig1}. Here, $B_i$ corresponds to the inner structuring element denoted by the region $EFGH$, while $B_o$ is denoted by the region $ABCD$. Additionally, $B_b$ is a structuring element that represents region $MNOP$ and is positioned between $ABCD$ and $EFGH$. The symbol $\vartriangle B$ denotes the annular structure, which is seen as the darkened region in Figure \ref{fig1}. The modified white top-hat transformation can be mathematically expressed as the difference between the image and the opening of the picture with respect to the structuring element $B_{oi}$. The detection of small targets is improved with the implementation of this modified technique, which effectively leverages the contrast in grey intensity between the target region and the surrounding cluttered background.

\begin{equation}
	 \hat{T_w} (f) =f-f \odot \hat b \label{eq3}
\end{equation}
where,
\begin{equation}
f \odot \hat b =\left(  f \ominus {\vartriangle B} \right) \oplus  B_b \label{eq4}
\end{equation}
The MTHM method is commonly accompanied by the use of a global threshold to distinguish between target and background pixels in order to achieve segmentation of the detection plane. This step is of paramount importance as it plays a critical role in distinguishing the target pixels that have been enhanced through MTHM from the background pixels. By establishing an optimal threshold value, it becomes feasible to distinguish the regions that potentially contain small targets from the complex background, which greatly facilitates the effective detection and analysis of these targets.

\par
\begin{figure}[t]
	\centerline{\includegraphics[scale=0.8]{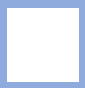}}
	\caption{Structuring element for CM.}
	\label{fig4}
\end{figure}

 Contour Morphology (CM) \cite{bai2008new, bai2010enhancement}  is a distinct iteration of Modified Top-Hat Morphology, characterized by a series of image operations. The procedure associated with Contour Morphology commences with the application of a dilation operation, followed by an erosion operation, to the input image $f$. This sequence of operations is mathematically represented as shown in Equation \ref{eq5}, and it plays a vital role in amplifying the detection of particular features or objects within the image. Contour Morphology serves as an image processing technique designed to emphasize specific attributes or traits in the image, rendering them more conspicuous and contributing to the identification of particular elements, such as small targets. 
\begin{equation}
f\circledcirc b=\left(  f \oplus  b_1 \right) \ominus b_2 \label{eq5}
\end{equation}
In the realm of Contour Morphology (CM), specific mathematical symbols and operations are employed: $\oplus$ represents the dilation operation, $\ominus$ stands for the erosion operation, and $\circledcirc$ signifies the contour morphology operation. CM utilizes two distinct structuring elements: $b_1$ takes the form of a one-pixel wide ring structure, illustrated in Figure \ref{fig4}, while $b_2$ is a square-shaped element. The background estimation, denoted as $f\circledcirc b$, is calculated using the contour morphology operation. This estimated background is then subtracted from the original scene image $f$ to generate the detection plane $DP$, as expressed in Equation \ref{eq6}. This process effectively enhances the detection of specific features or elements in the image, such as small targets, by emphasizing their contours and boundaries.

\begin{equation}
DP = f- f\circledcirc b \label{eq6}
\end{equation}
Following the application of the contour morphology operation and the subsequent generation of the detection plane (DP), a universal threshold is utilized to divide this DP into two groups of pixels: those representing potential target elements and those representing the background. This thresholding procedure aids in the differentiation of regions likely to encompass small targets from the neighboring background, simplifying the identification and analysis of the specific elements of interest in the image.

\par The Method of Directional Derivatives (MODD) \cite{bai2008new} is another approach in the category of image processing-based techniques for the detection of small targets. MODD operates on the assumption that a specific target within an image exhibits relatively isotropic behavior compared to the surrounding clutter and displays a significant magnitude of directional derivative in multiple directions. This technique is designed to enhance the visibility of the target while rejecting the influence of clutter within a scene image. The process involves a series of steps to achieve this goal
The directional derivatives of the scene are computed up to the first and second order in four distinct directions. The values of derivatives at $0^{\circ}, 45^{\circ}, {-}45^{\circ}, 90^{\circ}$ may be expressed using the Equations \ref{modd1} and \ref{modd2}.
\begin{equation}
{d}_{\alpha}^1 = ({K}_2 - \frac{17}{5}{K}_7-2{K}_9)\sin{\alpha}-({K}_3 - \frac{17}{5}{K}_{10}-2{K}_8)\cos{\alpha}\label{modd1}
\end{equation}

\begin{equation}
{d}_{\alpha}^2 = 2{K}_4\sin^2{\alpha} + 2{K}_5\sin{\alpha}\cos{\alpha} + 2{K}_6 \cos^2{\alpha}\label{modd2}
\end{equation}
The matrices $K_2 - K_{10}$ are derived from convolving the scene image with $5 \times 5$ matrices $W_2 - W_{10}$, as provided
\[ 
W_2 = \frac{1}{50} \begin{bmatrix} 2 & 2 & 2 & 2 & 2\\ 1 & 1 & 1 & 1 & 1 \\ 0 & 0 & 0 & 0 & 0 \\ -1 & -1 & -1 & -1 & -1\\-2 & -2 & -2 & -2 & -2
\end{bmatrix}\]

\[
 W_3 = \frac{1}{50} \begin{bmatrix} 2 & 1 & 0 & -1 & -2\\ 2 & 1 & 0 & -1 & -2 \\ 2 & 1 & 0 & -1 & -2 \\ 2 & 1 & 0 & -1 & -2\\2 & 1 & 0 & -1 & -2
\end{bmatrix} \]
\[ W_4 = \frac{1}{70} \begin{bmatrix} 2 & 2 & 2 & 2 & 2\\ -1 & -1 & -1 & -1 & -1 \\ -2 & -2 & -2 & -2 & -2 \\ -1 & -1 & -1 & -1 & -1\\2 & 2 & 2 & 2 & 2
\end{bmatrix}\]
\[ W_5 = \frac{1}{100} \begin{bmatrix} 4 & 2 & 0 & -2 & -4\\ 2 & 1 & 0 & -1 & -2 \\ 0 & 0 & 0 & 0 & 0 \\ -2 & -1 & 0 & 1 & 2\\-4 & -2 & 0 & 2 & 4
\end{bmatrix} \]
\[ W_6 = \frac{1}{70} \begin{bmatrix} 2 & -1 & -2 & -1 & 2\\2 & -1 & -2 & -1 & 2\\2 & -1 & -2 & -1 & 2\\2 & -1 & -2 & -1 & 2\\2 & -1 & -2 & -1 & 2
\end{bmatrix} \]
\[ W_7 = \frac{1}{60} \begin{bmatrix} 1 & 1 & 1 & 1 & 1\\ -2 & -2 & -2 & -2 & -2 \\0 & 0 & 0 & 0 & 0  \\ 2 & 2 & 2 & 2 & 2\\-1 & -1 & -1 & -1 & -1
\end{bmatrix} \]
\[ W_8 = \frac{1}{140} \begin{bmatrix} 4 & 2 & 0 & -2 & -4\\ -2 & -1 & 0 & 1 & 2 \\-4 & -2 & 0 & 2 & 4  \\ -2 & -1 & 0 & 1 & 2\\4 & 2 & 0 & -2 & -4
\end{bmatrix} \]
\[ W_9 = \frac{1}{140} \begin{bmatrix} 4 & -2 & -4 & -2 & 4\\ 2 & -1 & -2 & -1 & 2 \\0 & 0 & 0 & 0 & 0  \\ -2 & 1 & 2 & 1 & -2\\-4 & 2 & 4 & 2 & -4
\end{bmatrix} \]
\[ W_{10} = \frac{1}{60} \begin{bmatrix} 1 & -2 & 0 & 2 & -1\\ 1 & -2 & 0 & 2 & -1\\1 & -2 & 0 & 2 & -1\\1 & -2 & 0 & 2 & -1\\1 & -2 & 0 & 2 & -1
\end{bmatrix} \]
After applying this enhancement procedure, the scene image's directional derivatives are subjected to convolution with E-filters. Each filter, denoted as $E^1_{\alpha}, E^2_{\alpha}$, is generated by deriving the first and second-order directional derivatives of a Gaussian function in four directions: $\alpha = 0^{\circ}, 45^{\circ}, {-}45^{\circ}, 90^{\circ}$. The resulting enhanced images, represented as $f^1_{\alpha}$ and $f^2_{\alpha}$, are obtained using following Equation \ref{modd3}.
\begin{equation}
f^1_{\alpha} = d^1_{\alpha} \ast E^1_{\alpha}\hspace{0.5cm} and \hspace{0.5cm} f^2_{\alpha} = d^2_{\alpha} \ast E^2_{\alpha} \label{modd3}
\end{equation}
where the E-filters are given as
\[E_0^1 =  \begin{bmatrix} −0.1884 & 0 & 0.1884\\ −0.1991 & 0 & 0.1991 \\ −0.1884 & 0 & 0.1884
\end{bmatrix}\]
\[E_0^2 =  \begin{bmatrix} −0.0381 & −0.0762 & −0.0381\\ −0.0381 & −0.0762 & −0.0381 \\ −0.0381 & −0.0762 & −0.0381 
\end{bmatrix} \]
\[E_{45}^1 =  \begin{bmatrix} −0.2664 & −0.1408 & 0\\ −0.1991 & 0 & 0.1991 \\ −0.1884 & 0 & 0.1884
\end{bmatrix}\]
\[E_{45}^2=  \begin{bmatrix} −0.0306 & −0.0571 & −0.0456\\ −0.0571 & −0.0762 & −0.0571 \\ −0.0456 & −0.0571 & −0.0306
\end{bmatrix}\]
\[E_{-45}^1=  \begin{bmatrix} 0 & 0.1408 & 0.2664\\ −0.1408 & 0 & 0.1408 \\−0.2664 & −0.1448 & 0
\end{bmatrix}\]
\[E_{-45}^2=  \begin{bmatrix} −0.0456 & −0.0571 & −0.0306\\ −0.0571 & −0.0762 & −0.0571 \\ −0.0306 & −0.0571 & −0.0456
\end{bmatrix}\]
\[E_{90}^1 =  \begin{bmatrix} −0.1884 & −0.1991 & −0.1884\\ 0 & 0 & 0 \\ 0.1884 & 0.1991 & 0.1884
\end{bmatrix}\]
\[E_{90}^2=  \begin{bmatrix} −0.0381 & −0.0381 & −0.0381\\ −0.0762 & −0.0762 & −0.0762\\ −0.0381 & −0.0381 & −0.0381
\end{bmatrix}
\]
The objective of this phase is to enhance the detectability of targets that possess attributes consistent with the derivatives of a Gaussian function. After the construction of the Detection Plane (DP), it is crucial to acknowledge that the improved images $f^1_{\alpha}$ and $f^2_{\alpha}$ obtained in the preceding stage may exhibit negative values. To achieve this, all negative values are adjusted to zero, leading to the formation of the detection plane $DP$ as described in Equation \ref{modd:dp}.

\begin{equation}
DP = f_0^1 \star f_0^2 \star f_{45}^1 \star f_{45}^2 \star  f_{-45}^1 \star f_{-45}^2 \star f_{90}^1 \star f_{90}^2\label{modd:dp}
\end{equation}
These algorithms exploit the differences in frequency content between the intended target, the surrounding background, and extraneous noise. Typically, this frequency discrepancy is more pronounced and can be identified in the transform domain through the use of high-pass filters to remove the background and clutter noise. While frequency domain-based filtering detection algorithms demand more computational resources compared to spatial domain-based methods, advances in computer hardware have made frequency domain-based filtering algorithms increasingly practical in engineering applications. The primary high-pass filters used include the ideal high-pass filter, Gaussian high-pass filter, and Butterworth filter. Nevertheless, the former two filters exhibit some degree of the \textit{ringing} phenomenon, leading to incomplete filtering. In contrast, the Butterworth filter effectively addresses these issues. On this basis, researchers have explored alternative filtering techniques, such as the wavelet transform. This method is employed to separate high-frequency target data from low-frequency background data and enhance the image's signal-to-noise ratio through specific processing methods to achieve target detection.

In the early stages of developing traditional methods for detecting small and dim targets in IR imagery, researchers often encountered challenges related to the availability of suitable datasets. Consequently, many of these methods relied on self-constructed datasets. However, these self-generated datasets typically had limitations in terms of target diversity and variability, which made it challenging to establish robust benchmarks for evaluating algorithm performance. To tackle this issue, some researchers \cite{naraniya2021scene} have taken the initiative to create their own datasets for the purpose of performance evaluation. By generating datasets that cover a broader range of scenarios and target variations, their goal is to provide a more comprehensive assessment of algorithm performance and better reflect real-world conditions. This contributes to enhancing the credibility and reliability of evaluations of traditional small and dim target detection methods.

\par Nikhil et al. \cite{kumar2021detection,10249124} conducted an experiment in which they recorded a video sequence of the sky with clouds as the background using a panning IR camera. In this video, a small target with dimensions of 3 x 3 pixels and following a predefined trajectory was inserted into the image frames. The detected targets were then compared to the ground truth, and the counts of true positives and false alarms were recorded for each video frame. To assess the performance of the filtering-based methods mentioned earlier, they calculated the DR and FAR, which were averaged over 1000 frames.
\begin{figure}[t]
	\centerline{\includegraphics[scale=0.6]{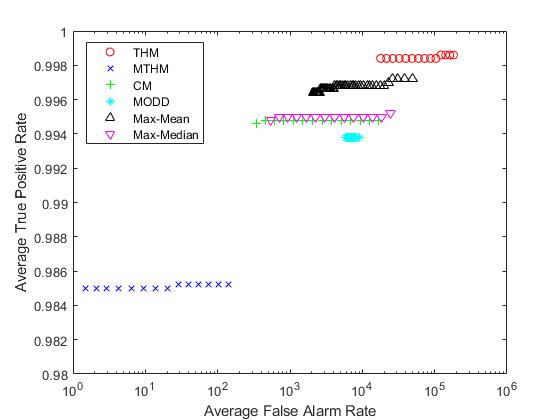}}
	\caption{Performance comparison of different filtering-based point target detection algorithms under varying segmentation thresholds, specifically focusing on scenarios with a less cluttered background.}
	\label{fig:comp}
\end{figure}

\begin{table*}[t]
\caption{Performance comparison of various filtering-based point target detection algorithms.}
\label{tab:filt}
\begin{tabular}{@{}cccccc@{}}
\toprule
\textbf{Reference} & \textbf{Method}       & \textbf{Type}                     & \textbf{Dataset}        & \multicolumn{2}{c}{\textbf{Result}} \\ \midrule
\multirow{2}{*}{\cite{gonzalez2004digital}}  & \multirow{2}{*}{THM}  & \multirow{2}{*}{Morphology Based} & \multirow{2}{*}{Custom \cite{10249124, kumar2021detection}} & Pd             & 0.9984             \\
                   &                       &                                   &                         & Fa             & 17942              \\ \\ \midrule
\multirow{2}{*}{\cite{bai2008new, bai2010enhancement}}  & \multirow{2}{*}{MTHM} & \multirow{2}{*}{Morphology Based} & \multirow{2}{*}{Custom \cite{10249124, kumar2021detection}} & Pd             & 0.981              \\
                   &                       &                                   &                         & Fa             & 1.501              \\ \\ \midrule
\multirow{2}{*}{\cite{bai2008new, bai2010enhancement}}  & \multirow{2}{*}{CM}   & \multirow{2}{*}{Morphology Based} & \multirow{2}{*}{Custom \cite{10249124, kumar2021detection}} & Pd             & 0.9946             \\
                   &                       &                                   &                         & Fa             & 341.52             \\ \\ \midrule
\multirow{2}{*}{\cite{gu2010kernel}} & \multirow{2}{*}{Max-mean}   & \multirow{2}{*}{Image Statistics Based} & \multirow{2}{*}{Custom \cite{10249124, kumar2021detection}} & Pd & 0.9964 \\
                   &                       &                                   &                         & Fa             & 2075               \\ \\  \midrule
\multirow{2}{*}{\cite{deshpande1999max}} & \multirow{2}{*}{Max-median} & \multirow{2}{*}{Image Statistics Based} & \multirow{2}{*}{Custom \cite{10249124, kumar2021detection}} & Pd & 0.9948 \\
                   &                       &                                   &                         & Fa             & 540.33             \\ \\ \midrule
\multirow{2}{*}{\cite{bai2008new}}  & \multirow{2}{*}{MODD} & \multirow{2}{*}{Gradient Based}   & \multirow{2}{*}{Custom \cite{10249124, kumar2021detection}} & Pd             & 0.9938             \\
                   &                       &                                   &                         & Fa             & 6034.66            \\ \\ \bottomrule 
\end{tabular}
\end{table*}

As shown in  Figure \ref{fig:comp} and Table \ref{tab:filt}, a comparison of various filtering-based methods for small and dim target detection in IR imagery reveals different trade-offs in terms of DR and FAR. The MTHM method tends to have the lowest FAR among the compared methods, indicating that it produces fewer false alarms. However, it comes at the cost of a lower DR, suggesting that it may miss some true targets.
THM demonstrates superior performance in terms of DR compared to other methods, meaning it is more effective at detecting true targets. However, it also results in a higher FAR, which implies it may produce more false alarms. CM and Max-Median methods exhibit similar performance metrics, with both DR and FAR falling in between the extremes observed with MTHM and THM. They offer a balance between detection and FARs. MODD appears to have a lower DR and FAR compared to other methods. It is a more conservative approach that may miss some targets but is less sensitive to changes in the threshold value. The choice of which method to use depends on the specific application's requirements and priorities. For example, in scenarios where minimizing false alarms is critical, MTHM might be preferred. On the other hand, if maximizing target detection is the primary goal, THM might be more suitable. Researchers and practitioners must weigh the trade-offs and select the method that best aligns with their operational needs. It is important to note that the dataset utilized in these works is tailored to include exclusively aerial scenarios acquired by a sensor installed on an aerial platform. It is worth highlighting that such scenarios typically exhibit significantly lower levels of clutter compared to ground scenarios captured from an aerial perspective.
\par Filtering-based methods are limited to reducing uniform background clutters and are unable to effectively reduce complex background noises. This limitation leads to high rates of false alarms and unstable performance.  In defense applications, it is necessary to handle numerous real-world scenarios that involve a significant amount of background clutter. The utilization of these methods may necessitate the implementation of supplementary algorithms for managing clutter and result in an escalation of computational expenses. 
Nevertheless, the aforementioned techniques relying on image processing, filtering, or manually designed features exhibit limited efficacy when confronted with complex scenarios such as targets exhibiting diverse shapes and sizes and backgrounds containing excessive clutter and noise.
In contrast, deep neural networks have the ability to autonomously learn complex features from extensive datasets that encompass intricate scenes, thanks to their end-to-end learning approach.
Traditional image processing methods often require hyper-parameter tuning, which can be a challenging issue. In defense applications, for instance, if one intends to apply such algorithms with automated systems like IRST or MAWS, the lack of human intervention makes it nearly impossible to adjust hyper-parameters according to different scenarios. 
Assuming the selection of the MTHM algorithm for an automated defense system, there exist three adjustable hyper-parameters: the diameter of the inner ring, the diameter of the outer ring, and the threshold for segmentation of the detection plane. Achieving robust performance across all scenarios poses a challenging task when tuning these three hyper-parameters simultaneously.

\subsection{Deep Learning Based Approaches}

The use of deep learning algorithms to detect small targets in IR images has proven to be significantly better than traditional methods. Traditional methods for target detection often rely on manual or rule-based approaches, which can be time-consuming, subjective, and less effective in complex scenarios. However, machine learning algorithms, particularly deep learning models, have revolutionized the domain of point target detection in IR imagery. One of the key advantages of machine learning algorithms is their ability to learn and adapt to large volumes of data. In the context of IR imagery, these algorithms can be trained on diverse datasets consisting of labeled images that contain small targets. By exposing the algorithms to a wide range of target characteristics, such as size, shape, orientation, and IR signatures, they can effectively learn the discriminative features that distinguish targets from background clutter.

Deep learning models, such as CNNs, have shown exceptional performance in detecting small targets in IR images. CNNs are specifically designed to automatically extract hierarchical features from images, enabling them to capture complex patterns and structures. They can learn to recognize distinctive IR signatures and subtle spatial arrangements associated with small targets, even in challenging conditions like low contrast or high background noise. Notwithstanding the considerable achievements of CNNs in object detection and segmentation \cite{zhang2022resnest}, there has been limited exploration of deep learning methodologies in the domain of IR small target detection. 
 Deep learning models require large amounts of data for training.  These datasets should ideally have high-quality annotations, which would enable researchers to develop, evaluate and compare new approaches for this task.
IR small targets frequently encounter the challenge of being immersed in complex backgrounds characterized by low signal-to-clutter ratios. In the context of networks, the task of identifying dim targets while minimizing false alarms requires a combination of a comprehensive understanding of the entire IR image at a higher level and a detailed prediction map with fine resolution. However, this poses a challenge for deep networks as they tend to prioritize learning semantic representations by gradually reducing the size of features \cite{he2016deep}.

Deep learning-based SIRST detection methods can be broadly classified into two categories. The initial category within this classification encompasses methodologies that heavily depend on complete supervision. In recent years, there has been a significant increase in research interest in methods that rely on fully supervised learning. However, these methods are associated with high annotation costs due to the requirement of a large number of per-pixel annotations. The second category of this classification is based on the concept of Weak Supervision.  Under such approaches, point-level supervision is employed in the context of IR small target detection. Compared to earlier methods, these approaches include a lower annotation cost for per-pixel annotation.

\subsubsection{Fully Supervised Learning Based Approaches}
Approaches under this category can be further classified into two subclasses: Detection type approaches and Segmentation type approaches. Detection-type approaches employ a generic framework for the purpose of detecting small and dim targets. In contrast, segmentation methods focus on the binary classification of pixels, distinguishing between foreground and background.

\paragraph{Detection Based Approaches}

Within this particular group of techniques, generic deep learning frameworks for detecting different types of targets are modified and adapted specifically for the purpose of detecting small and dim point targets. Liu et al. \cite{liu2017image} offered a deep learning-based end-to-end solution for small target detection that might be viewed as a classifier approach. In comparison to traditional image processing-based methods, experimental results show that this method is robust and insensitive to background and altering targets. The network is constructed with an input dimension of 21×21 pixels. Small input neural networks are utilized as a mobile filter window for the purpose of detecting small targets located at any position within an image. In the training phase, the only pre-processing step involves subtraction of the mean value, which is calculated on the training set, from every pixel. The image undergoes a process of layering. Each layer is fully connected to the following layer. The initial layers consist of 128 channels each, while the final layer executes binary classification. The final layer is the soft-max transformation. All hidden layers are equipped with rectification non-linearity. Details of different trained models labeled as A, B, C, D, and E can be seen in Table \ref{tab:mlp}. 

\begin{table*}[t]
\caption{MLP-based architectures for small and dim target detection.}
\label{tab:mlp}
\resizebox{2\columnwidth}{!}{%
\begin{tabular}{@{}lccccc@{}}
\toprule
\multicolumn{1}{c}{\textbf{Layer Type}}                                     & \textbf{Model A}      & \textbf{Model B}      & \textbf{Model C}      & \textbf{Model D}      & \textbf{Model E}      \\ \midrule
Input Layer   & - & 441X128 + RELU & 441X128 + RELU   & 441X128 + RELU   & 441X128 + RELU   \\
Middle Layers & - & -              & 1X128X128 + RELU & 2X128X128 + RELU & 3X128X128 + RELU \\
Output Layer                                                                & 441X2 + Softmax & 128X2 + Softmax & 128X2 + Softmax & 128X2 + Softmax & 128X2 + Softmax \\
\begin{tabular}[c]{@{}l@{}}FC  Layers\end{tabular} & 1               & 2               & 3               & 4               & 5               \\ \bottomrule
\end{tabular}
}
\end{table*}

Furthermore, a number of cutting-edge networks \cite{hu2018squeeze, li2019selective} are specifically engineered for generic image datasets. Utilizing them directly for IR small target detection can result in severe failure due to the significant disparity in the data distribution. It necessitates the reconfiguration of the network across various dimensions.
 Numerous studies highlight the importance of aligning the receptive fields of CNNs with the scale range of objects being analyzed \cite{li2019scale,  singh2018analysis}. Without modifying the down-sampling method, it becomes increasingly difficult to maintain the ability to detect small IR targets as the network becomes deeper.
 Current attention modules commonly aggregate global or long-range contexts \cite{fu2019dual, hu2018squeeze}. The underlying assumption is that the objects being referred to are of significant size and have a wide distribution. However, this is not applicable to IR small targets, as a global attention module would diminish their distinctive characteristics. This prompts the inquiry of which type of attention module is appropriate for detecting IR small targets.
 Recent studies have incorporated cross-layer features in a unidirectional top-down approach \cite{li2018pyramid, wang2019salient}, with the goal of choosing appropriate low-level features based on high-level semantics. However, due to the potential for small targets to be obscured by background noise in deeper layers, relying solely on top-down modulation may not be effective and could potentially have negative consequences.

\par
\begin{figure}[]
	\centering
	\includegraphics[scale=0.8]{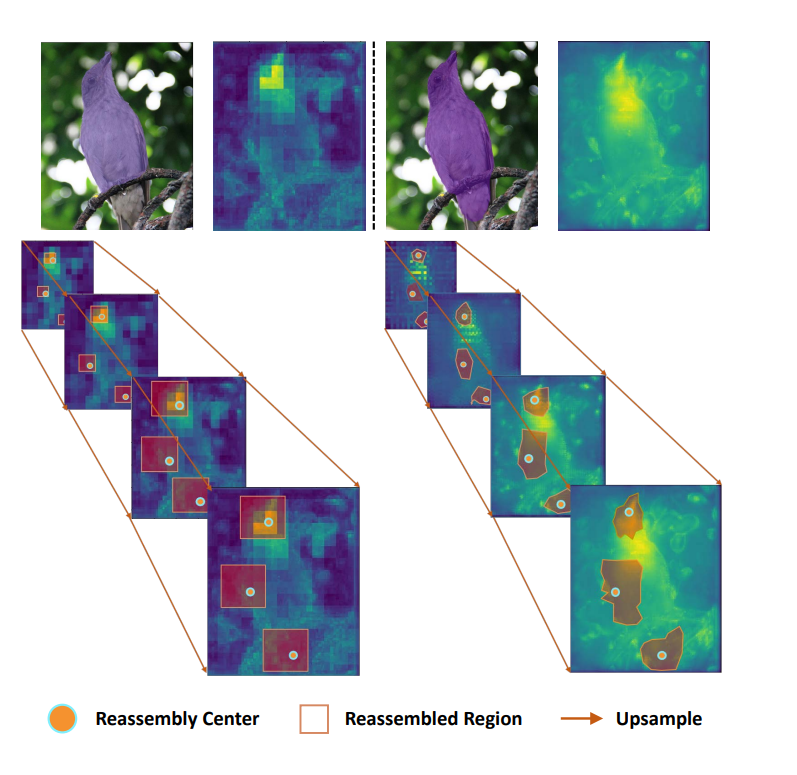}
	\caption{Illustration of CARAFE working mechanism. Figure courtesy \cite{wang2019carafe}.}
	\label{carafe}
\end{figure}

 In an effort to address the challenge of losing critical features of small IR targets during the network sampling process, Mou et al. \cite{mou2023yolo} introduced YOLO-FR, a model designed for the detection of small targets in IR images. YOLO-FR is based on the YOLOv5 \cite{redmon2016you, redmon2018yolov3, bochkovskiy2020yolov4, ge2021yolox, wang2023yolov7} architecture and integrates feature reassembly sampling techniques, allowing for resizing the feature map while preserving existing feature information. To prevent feature loss during down-sampling, a Spatial Temporal Down-sampling (STD) Block is employed, which retains spatial information within the channel dimension. Additionally, the CARAFE \cite{wang2019carafe} operator is utilized to expand the feature map size without distorting the feature mapping mean. To facilitate the down-sampling process, an STD block is specifically designed to reduce image resolution, effectively transferring additional spatial domain information to the depth dimension to enhance the extraction of small target features without introducing parameter inflation. This block is responsible for all down-sampling operations within the backbone network.
As shown in Figure \ref{carafe}, the up-sampling process in the feature fusion network employs the CARAFE operator, which is a region-content-based up-sampling technique involving two key steps: the prediction of up-sampling kernels and their application to the original map positions. Evaluation metrics and visualization results indicate a substantial improvement in the model's ability to detect small targets following the incorporation of the CARAFE operator. To enhance the detection of small targets using shallow detailed features, the feature fusion network has been extended for features extracted from the backbone network after down-sampling for fusion. Additionally, the authors incorporated a small target detection head with a reduced receptive field. The authors conducted experiments to determine the optimal combination of target detection heads. The dataset utilized in their research was the publicly accessible infrared dim-small aeroplane target dataset provided by Liu et al. \cite{hui2020dataset}. The dataset consisted of 22 sets of data, comprising a total of 16,177 infrared images. Each image had dimensions of 256 × 256 pixels.

\paragraph{Segmentation Based Approaches}

These approaches utilize a deep learning framework to conduct binary classification of scenes, distinguishing between foreground and background. Typically, small and dim targets are categorized as foreground entities, while the remaining elements are designated as background components.

\begin{figure}[t]
	\centerline{\includegraphics[scale=1]{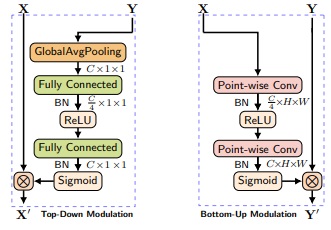}}
	\caption{Portrayal of one-directional modulation modules: Top-down global attentional modulation as well as bottom-up point-wise attentional modulation. Figure courtesy \cite{dai2021asymmetric}.}
	\label{fig:ACM}
\end{figure}
\par Dai et al. \cite{dai2021asymmetric} introduced the Asymmetric Contextual Modulation Network (ACMnet). ACMnet presents a network architecture that places a strong emphasis on customizing the down-sampling process, attention mechanisms, and feature fusion techniques. The primary goal is to effectively preserve the features related to small IR targets. It is essential that the network's receptive fields of predictors are adapted to match the scale range of the objects to ensure the preservation of IR small target features as the network delves deeper. Failure to customize the down-sampling approach could result in the loss of vital information. ACMnet effectively handles this challenge by carefully adjusting the down-sampling process to ensure the capture and retention of pertinent features of IR small targets. Traditional attention modules are typically designed to aggregate global or long-range contextual information, assuming that objects are large and distributed globally. However, this assumption does not hold for IR small targets, where a global attention module can potentially weaken their features. ACMnet introduces an attention module that specifically aims to enhance the visibility of IR small targets. By employing a more localized attention mechanism, ACMnet enhances its detection capabilities for these specific targets.

Recent methodologies incorporate cross-layer feature fusion in a top-down manner, selecting low-level features based on high-level semantic information. However, the presence of background interference in deep layers can overshadow small targets, rendering the pure top-down modulation ineffective or even detrimental. ACMnet overcomes this limitation by redesigning the feature fusion approach, incorporating mechanisms to account for the potential interference of small targets by background noise. This approach offers an innovative solution to address the challenges arising from the size discrepancy between IR small targets and objects in general datasets. The solution involves integrating the ACM mechanism as a plug-in module into various host networks, enabling the bidirectional transfer of abstract concepts and specific implementation details across different feature levels. As shown in Figure \ref{fig:ACM}, this approach enhances the efficiency of small target detection in IR imagery by incorporating a top-down pathway that incorporates high-level semantic feedback and a bottom-up pathway that encodes finer visual details into deeper layers. This is made possible by utilizing Global Channel Attention Modulation (GCAM) for top-down modulation and Pixel-wise Channel Attention Modulation (PCAM) for bottom-up modulation. PCAM is specifically designed to enhance and maintain the visibility of IR small targets. The ACM module replaces the existing cross-layer feature fusion operations, resulting in improved network performance with only a minimal increase in the number of parameters. To emphasize the subtle details of IR small targets in deep layers, the authors have introduced a point-wise channel attention modulation module, aggregating the channel feature context for each spatial position individually. Unlike top-down modulation, this modulation pathway propagates context information in a bottom-up manner to enrich high-level features with spatial details from low-level feature maps.

\begin{figure*}[]
	\centering
	\includegraphics[scale=1.2]{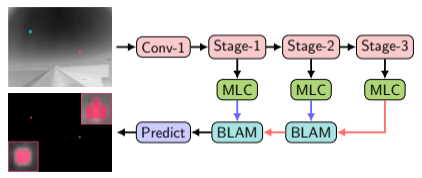}
	\caption{Illustration of ALCNet archetecture, as suggested, integrates same-layer multi-scale local contrast (MLC) modules and cross-layer bottom-up local attentional feature modulation (BLAM) modules into a feature pyramid network. The blue line signifies the channel number transformation, while the red line denotes the upsampling operator. Figure courtesy \cite{dai2021attentional}.}
	\label{fig:alc}
\end{figure*}
\par Attentional Local Contrast Network (ALCNet) \cite{dai2021attentional}, as shown in Figure \ref{fig:alc},  introduced an innovative approach to address the challenge of detecting IR small targets in a single image by integrating the feature learning capabilities of deep networks and the physical mechanisms of model-driven methods into an end-to-end network. ALCNet is specifically designed for the detection of single-frame IR small targets, and it brings two substantial enhancements. Firstly, the authors introduced an acceleration strategy that utilizes feature map cyclic shifts, modularizing a local contrast measure method developed by Wei et al. \cite{wei2016multiscale}. The modularization is achieved through the creation of a depth-wise parameter-less nonlinear feature refinement layer, which holds a clear physical interpretation and addresses the limited receptive field imposed by convolutional kernels. This refinement layer encodes longer-range contextual interactions. Additionally, the network's downsampling approach has been adjusted to enhance and retain the characteristics of small targets. They introduced a Bottom-up Attentional Modulation module, which encodes finer details from low-level features into the higher-level features of deeper layers. The feature maps obtained through cross-layer fusion are utilized for segmentation purposes. A specific model-inspired module is employed to encode the input image into local contrast measures. This approach effectively combines both labeled data and domain knowledge to leverage the full capacity of the network. Consequently, the network's ability to autonomously learn discriminative features overcomes the limitations of inaccurate modeling and sensitivity to hyperparameters often encountered in model-driven methods. Furthermore, it tackles the difficulty of having minimal intrinsic features in data-driven methods by incorporating the domain knowledge of the local contrast prior into deep neural networks. 

ALCNet \cite{dai2021attentional} utilizes a local contrast foundation, measurements of local contrast at multiple scales, and bottom-up attentional adjustment to boost its detection performance. The network is trained using a modified ResNet-20 backbone that serves as a feature extractor, capturing high-level semantic features from the input image. To address the class imbalance between small targets and the background, the Soft-IoU loss function is employed to optimize the network's segmentation task. These strategies significantly contribute to the improved performance of ALCNet in small target detection. By highlighting and preserving crucial small target features, the network improves its ability to differentiate targets from the background. The authors also provided comprehensive ablation studies to evaluate the effectiveness and efficiency of the network architecture. In contrast to methods that rely solely on either data \cite{zhao2019tbc} or models \cite{wang2017infrared}, this approach maximizes the integration of both labeled data and domain knowledge. As a result, it effectively addresses the issues of inaccurate modeling and hyper-parameter sensitivity inherent in model-driven methods by enabling the network to autonomously learn discriminative features. Additionally, it mitigates the challenge of minimal intrinsic features in data-driven approaches by incorporating domain knowledge of the local contrast prior to deep networks. This approach underscores the potential of convolutional networks that integrate local contrast prior, a feature traditionally addressed in non-learning methods. ALCNet demonstrates promising outcomes in IR small target detection by breaking traditional constraints and fusing local contrast feature maps across layers. By uniting deep networks with domain knowledge, the network achieves enhanced accuracy and efficiency in detecting small IR targets, highlighting the importance of incorporating domain knowledge into deep learning architectures to tackle the challenges of small target detection in IR imagery.

\par The objective of SIRST detection is to differentiate small targets from complex backgrounds in IR images. While CNN-based methods have demonstrated promise in generic object detection, their direct application to IR small targets is hindered because the pooling layers within these networks can lead to the loss of target information in deeper layers. To tackle this challenge, the Dense Nested Attention Network (DNA-Net), as presented in the work by Li et al. \cite{li2022dense}, is introduced. DNA-Net draws inspiration from the success of nested structures in medical image segmentation \cite{zhang2018mdu} and hybrid attention in generic object detection \cite{woo2018cbam}. Several key innovations in DNA-Net contribute to its enhanced performance in the detection of IR small targets. Primarily, the Dense Nested Interactive Module (DNIM) in DNA-Net facilitates a gradual interaction between high-level and low-level features. This interaction allows information about small targets to propagate through the network's layers without any loss, enabling the preservation of crucial target details even in deep layers. By ensuring the presence of small targets throughout the network, DNA-Net effectively addresses the issue of target loss, a common problem in traditional CNN-based methods. In addition, the cascaded Channel and Spatial Attention Module (CSAM) further enhances features by adapting them to the specific characteristics of targets and cluttered backgrounds. CSAM employs attention mechanisms that selectively emphasize pertinent information while suppressing irrelevant noise, enhancing the network's discriminative power. This adaptive feature enhancement ensures that DNA-Net focuses on the most salient aspects of the targets, improving their visibility and aiding in accurate detection. The combination of DNIM and CSAM in DNA-Net results in a synergistic effect, with the network benefiting from both progressive feature fusion and adaptive enhancement. This allows the network to effectively incorporate and exploit contextual information, leading to superior performance in detecting IR small targets.

\begin{figure*}{}
	\centering
	\includegraphics[scale=1.5]{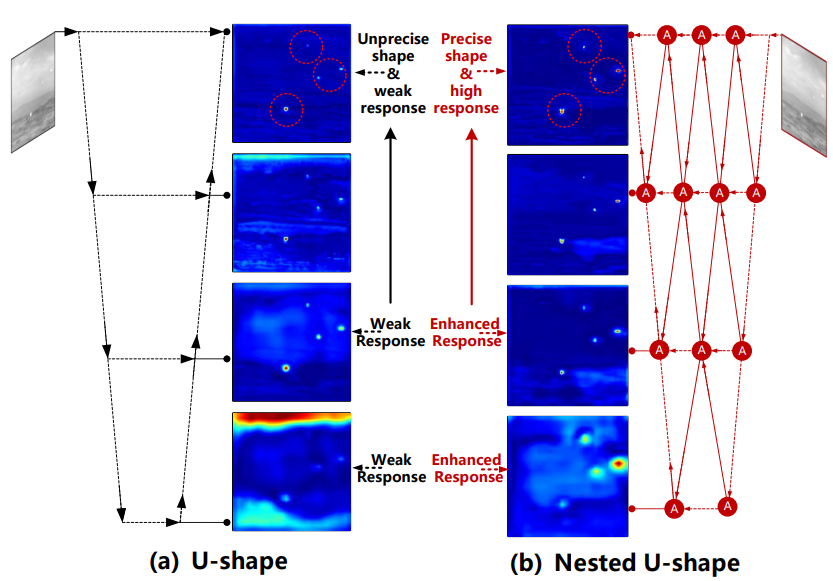}
	\caption{Portrayal of small targets within the deep convolutional neural network layers of two network architectures: (a) U-shape network and (b) Dense Nested U-Shape (DNA-Net) network. Figure courtesy \cite{li2022dense}.}
	\label{dnanet}
\end{figure*}

DNA-Net  \cite{li2022dense} leverages the advantages of dense nested structures and attention mechanisms, enabling it to capture intricate details and distinguish small targets from cluttered backgrounds, even in challenging scenarios with varying target sizes, shapes, and clutter conditions. The architecture of DNA-Net comprises three main modules: feature extraction, feature pyramid fusion, and eight-connected neighborhood clustering. These modules collaborate to generate the final detection results for SIRST images. In the feature extraction module, DNA-Net employs DNIM and CSAM. Input SIRST images undergo pre-processing and are passed through the DNIM backbone to extract multi-layer features. DNIM incorporates skip connections with intermediate convolution nodes to facilitate iterative feature fusion at different layers. To bridge the semantic gap that may arise during feature fusion, CSAM is employed to adaptively enhance these multi-level features, ensuring improved feature fusion and representation.
The feature pyramid fusion module focuses on the combination of enhanced multi-layer features from DNIM. These features are initially upscaled to the same size, ensuring consistency across scales. Subsequently, feature maps from shallow layers, rich in spatial information, and deep layers, rich in high-level semantic information, are concatenated to produce robust and comprehensive feature maps. The eight-connected neighborhood clustering module receives the feature maps from the previous stage and determines the spatial location of the target's centroid through calculations. The centroid serves as a reference point for later stages of comparison and evaluation. DNA-Net architecture follows a sequential process of feature extraction, feature pyramid fusion, and eight-connected neighborhood clustering to detect IR small targets effectively. By harnessing the capabilities of DNIM and CSAM, the network achieves progressive feature fusion, adaptive enhancement and robust representation of small targets, ultimately improving the accuracy and reliability of detection results. Figure \ref{dnanet} illustrates the presence of small targets within the layers of two network topologies: the U-shape network and the Dense Nested U-shape (DNA-Net) network.

\par The authors in their work \cite{liu2021infrared} have introduced a feature enhancement module, which operates as a feed-forward network. Its purpose is to obtain more distinct features associated with small-sized targets. Given the limited sizes and faint appearance of these targets, the failure to capture distinctive information from them significantly increases the chances of missing detections. Furthermore, the network may suffer from the potential loss of features related to small-sized targets. To mitigate this, an up-sampling structure akin to U-Net \cite{ronneberger2015u} has been integrated to enhance the retrieval of information concerning these compact targets. The authors have proposed an approach for the detection of small-sized targets. In this system, the self-attention mechanism borrowed from the transformer model is employed to gather interaction information among all embedded tokens. This capability allows the network to discern the differentiation between small-sized targets and the background within a broader context. This research marks one of the first endeavors to explore the use of transformers for the detection of small IR targets. The introduced feature enhancement module has significantly contributed to obtaining a more comprehensive understanding of discriminative features associated with targets. This innovative method consists of three main components; a feature embedding module, which is intended to extract a succinct feature representation of an image; a compound encoder, employed to gather information about the interactions between all embedded elements and to derive more distinct features for smaller targets; and a specialized decoder algorithm designed for producing confidence maps.

\par
\begin{figure*}{}
	\centering
	\includegraphics[scale=1]{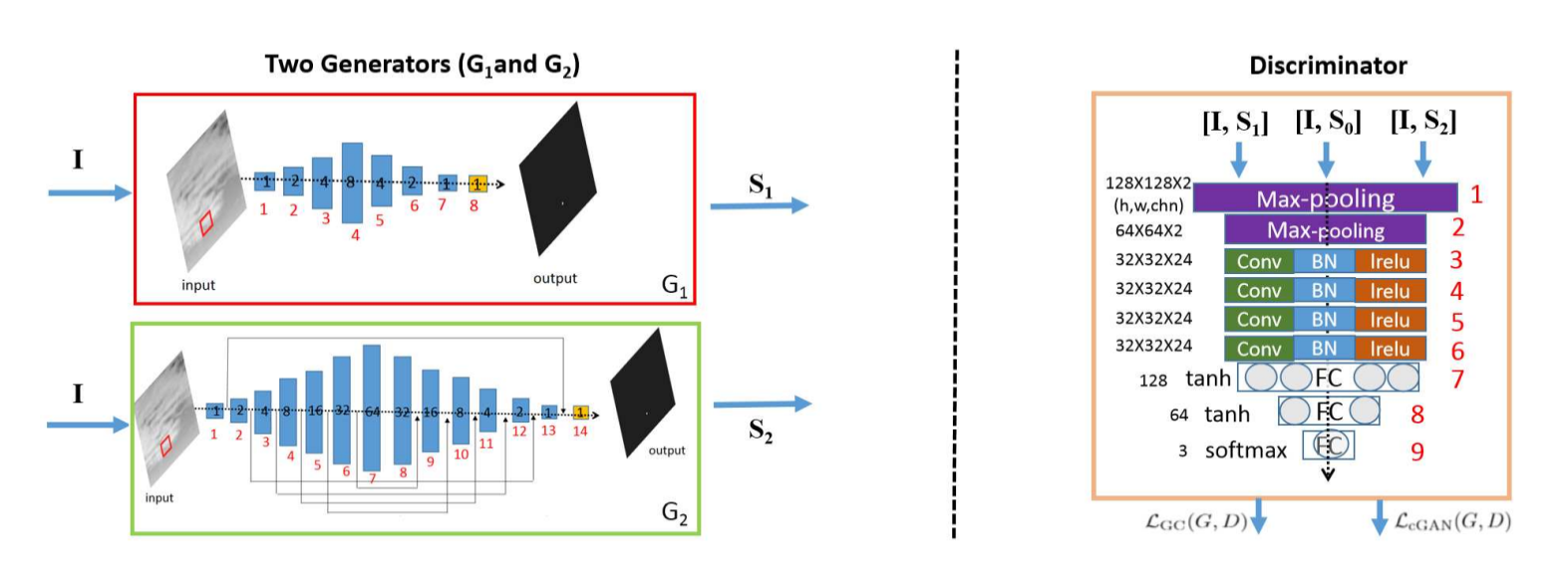}
	 \caption{An overview of the system and network architectures. There are two generators located on the left side. Generator One ($G_1$) is visually represented in the red box, whereas Generator Two ($G_2$) is depicted in the green box. The discriminator is depicted in the orange frame on the right. Red numerals represent the index of layers. The dilation factor utilized by a given generator layer is represented by the black number contained within that layer. The width, height, and channel count of feature maps in each layer are provided in addition to that layer for the discriminator. Figure courtesy \cite{wang2019miss}.}
	\label{msfa}
\end{figure*}

 The model proposed by \cite{wang2019miss}, as depicted in Figure \ref{msfa}, is structured around a conditional Generative Adversarial Network (GAN) architecture that includes two generator networks and one discriminator network. Each generator is specialized for a distinct sub-task, while the discriminator's role is to differentiate between the three segmentation outcomes produced by the two generators and the ground truth. Furthermore, to enhance their performance in handling sub-tasks, the two generators incorporate context aggregation networks with varying receptive field sizes. This enables them to encompass both local and global perspectives of objects during the segmentation process. The model is composed of generator and discriminator components, similar to the conditional Generative Adversarial Network (cGAN).

In contrast to the traditional cGAN, the model proposed by \cite{wang2019miss} features two generators, namely $G_1$ and $G_2$, and a single discriminator denoted as $D$. Each generator function takes an input image, denoted as $I$, and generates an output image representing the segmentation result. The primary objective of the generators is to minimize metrics like MD or FA during the segmentation process. To facilitate adversarial learning, the discriminator is specifically designed to differentiate between three segmentation outcomes: $S_1$, $S_2$, and $S_0$. Here, $S_0$ corresponds to the accurate segmentation of the ground truth, where "1" represents objects and "0" represents the background. While it's possible to train the two generator networks separately and then combine their segmentation results, this fusion-after-training approach limits the exchange of information during their training process, resulting in sub-optimal segmentation outcomes. To address this issue, they employ a cGAN framework to jointly train both generators. More precisely, it utilizes the discriminator $D$ as an intermediary to establish a connection between $G_1$ and $G_2$, thereby facilitating information exchange between the two generators. This information exchange has the potential to enhance the effectiveness of $G_1$, initially designed to minimize MD, in reducing FA and similarly enhance the effectiveness of $G_2$, initially designed to minimize FA, in reducing MD. Additionally, both generators receive robust supervision signals from $D$ due to the adversarial mechanism, which compels them to converge towards the ground truth to deceive $D$. Through this process, the two generators ultimately generate segmentation results that are consistent and closely resemble the ground truth. Once the entire model has been trained, either generator can be employed to process a test image and generate a segmentation result. This is possible because both generators have undergone training to reach convergence via the adversarial learning process.

\par Peng et al. \cite{peng2023dynamic} introduced the Dynamic Background Reconstruction (DBR) approach for detecting small and dim targets, which comprises three modules: the Detection Head (DH), the Background Reconstruction module (BR), and the Dynamic Shift Window (DSW). Initially, the DSW algorithm calculates an offset value based on the target object's ability to shift toward the patch's center in raw IR images. The BR algorithm then dynamically adjusts window positions based on this offset, utilizing the Naive Background Reconstruction (NBR) to restore clean backgrounds. To enhance detection performance, the approach combines IR images with targets, background-only images, and their differences before inputting them into the DH algorithm. The recall rate surpasses the precision rate due to the detector's tendency to classify reconstruction errors as positive events. The utilization of Vision Transformers in image processing, as described by Dosovitskiy et al. \cite{dosovitskiy2020image}, involves the partitioning of a given input image into patches of size 16x16. This approach provides inherent advantages in effectively dealing with mask tokens. MAE \cite{he2022masked} utilizes random patch removal to restore pixels when a high masking ratio is present.

DBR \cite{peng2023dynamic} effectively addresses the problem of a transformer model incorrectly dividing a target into adjacent patches, which can hinder background reconstruction. The authors introduced the DSW algorithm to calculate offsets for dynamic image shifting before input embedding. The DBR algorithm shows resilience against reconstruction errors. The authors proposed an approach incorporating the DH technique and the WDLoss method to mitigate the impact of reconstruction errors on detection performance by addressing specific aspects of the network architecture and loss function. When an IR image is initially fed into the DBR system, it undergoes processing with the DSW algorithm to determine offset values $(\Delta x, \Delta y)$. These offsets indicate the horizontal and vertical shifts necessary to reposition the target object to the center of a patch, avoiding division into multiple patches. Following the input embedding phase, the background reconstruction process involves subjecting the IR image to two masking processes that complement each other. This approach is commonly referred to as grid masking. The theoretical goal is to subtract the generated background from the original image. However, due to discrepancies between the generated background and the actual background, reconstruction errors occur. In order to minimize the influence of these errors on the performance of detection, the DH algorithm is employed to integrate the original image, the generated background, and their discrepancies.

\par
\begin{figure*}{}
	\centering
	\includegraphics[scale=0.8]{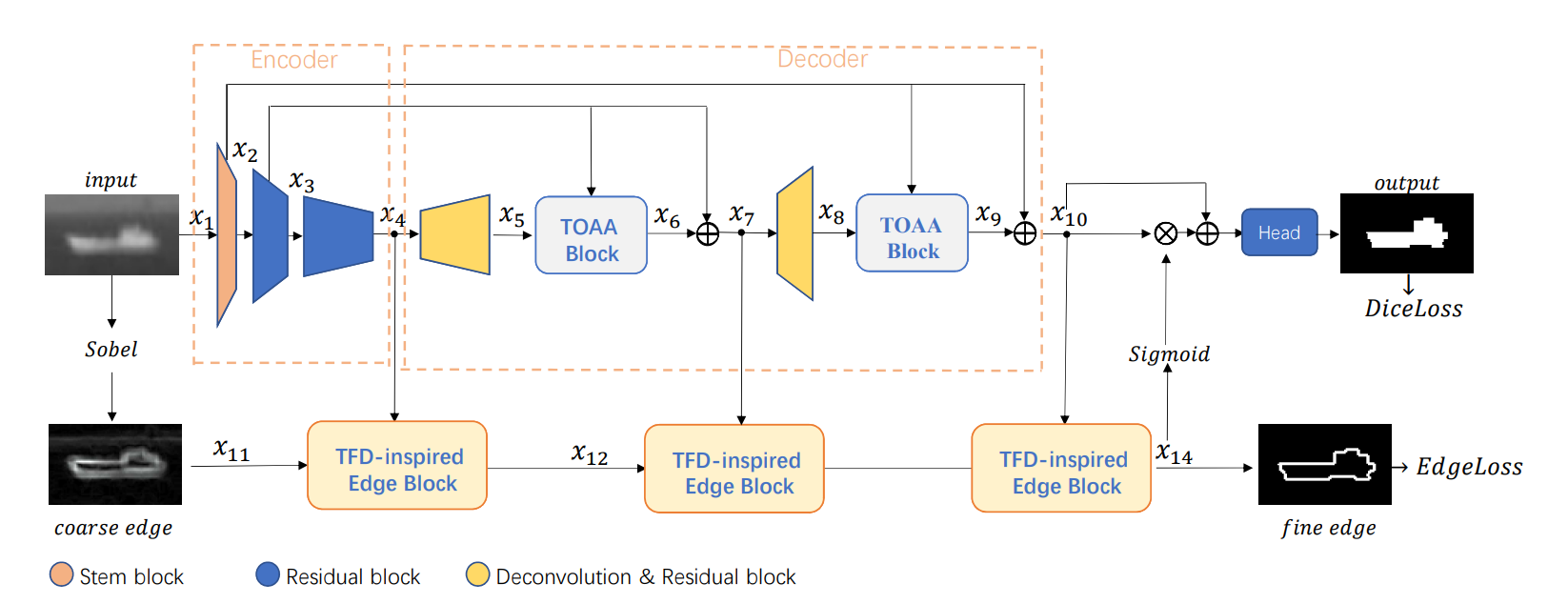}
	\caption{Portrayal of ISNet, which has a U-Net architecture featuring TOAA blocks and TFD-inspired edge blocks. Figure courtesy \cite{zhang2022isnet}.}
	\label{isnet}
\end{figure*}

 Precisely identifying shape details in the detection of IR small targets is a challenging task due to factors such as low SNR and poor contrast. These challenges often lead to targets getting obscured within noisy and cluttered backgrounds. To address this issue, the authors \cite{zhang2022isnet} presented an approach known as the IR Shape Network (ISNet)  in their work, as illustrated in Figure \ref{isnet}. ISNet comprises two key components: the Taylor Finite Difference (TFD)-inspired edge block and the Two-Orientation Attention Aggregation (TOAA) block.

The TFD-inspired edge block draws inspiration from the TFD algorithm, employing mathematical techniques to enhance edge information at various levels. This enhancement boosts the contrast between the target and the background, facilitating the extraction of shape information. The TFD-inspired edge block significantly contributes to improved target edge detection by combining information from different levels, thus enabling the network to effectively capture fine target edges. To address the issue of low-level features capturing intricate target details not found in high-level features, the authors introduced the TOAA block. This block incorporates two attention modules operating in parallel, generating attention maps in both row and column directions. These attention maps are utilized to adapt and enhance high-level features. Ultimately, the attentive features are aggregated and combined to produce the block's output. For training, the network employs two loss functions: the Dice loss  \cite{sudre2017generalised} and the Edge loss. The Dice loss quantifies the similarity between the predicted mask and the ground truth by comparing their intersection. Conversely, the Edge loss utilizes binary cross-entropy to assess the dissimilarity between the predicted mask and the ground truth concerning edge prediction. These two loss functions are weighted using a hyper-parameter referred to as lambda, and the final training objective combines the Edge loss and the Dice loss for mask prediction.

\par The performance of CNN-based IR small target detection is constrained by the small target size, complex backgrounds leading to clutter, and the inability of traditional CNNs to capture long-range dependencies. To overcome these challenges, the authors \cite{wang2022mpanet} introduced the Multi-Patch Attention Network (MPANet) in their work, which incorporates an axial-attention encoder and a Multi-Scale Patch Branch (MSPB) structure. The encoder architecture integrates axial attention mechanisms to improve the representation of small targets and reduce the impact of background noise. In order to mitigate the heavy computational cost associated with multi-head attention and facilitate the stacking of self-attention layers over wider regions, a factorization technique is employed to transform a two-dimensional self-attention into two separate one-dimensional self-attentions.

The axial attention mechanism, as proposed by Wang et al. \cite{wang2020axial}, incorporates relative positioning encodings for queries, keys and values, so augmenting the effectiveness of the model. Current methods suffer from significant loss of semantic information due to multiple pooling processes, which hinders their applicability to small IR target detection, primarily because of the low resolution of IR images. Additionally, the patch-wise training approach restricts the network's ability to capture mutual information or long-range dependencies between patches. To address these issues, an MSPB structure is used to perform aggregation operations hierarchically, beginning from the bottom and progressing upward. The global branch employs a position-sensitive attention block to capture semantic and contour features of both the foreground, background and target location information.
In the encoder, a lightweight U-shape architecture and axial attention block replace conventional convolutions. The structure of MSPB enables the effective extraction of feature representations at several semantic scales, including both coarse and fine-grained features. This is achieved by a bottom-up fusion process that integrates low-scale patch semantics with global picture semantic information. The experimental results on the publicly available SIRST dataset indicate that the MPANet demonstrates excellent performance in fine segmentation and target localization.

\begin{figure*}
	\centering
	\includegraphics[scale=0.9]{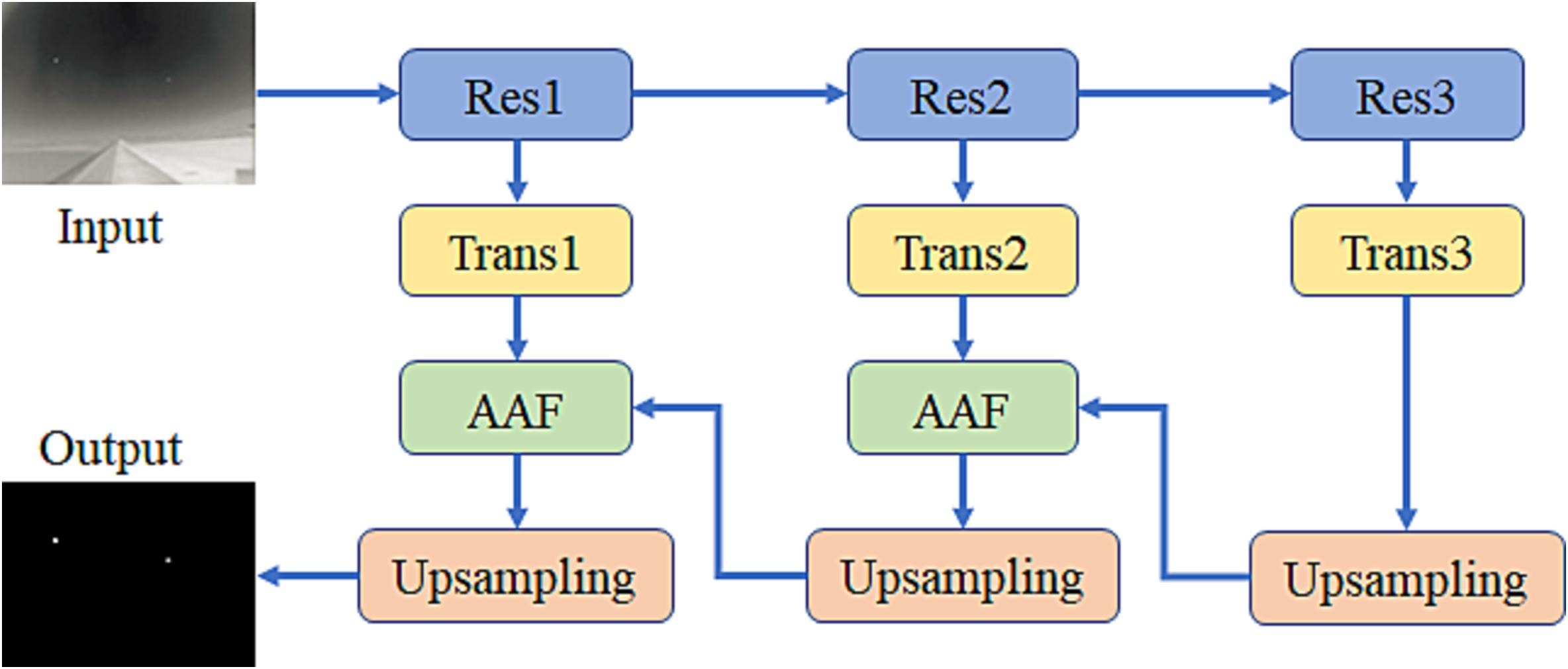}
	\caption{Architecture of the GANet framework. Figure courtesy \cite{zhang2024global}.}
	\label{GANet}
\end{figure*}

Zhang et al. \cite{zhang2024global} proposed a methodology for detecting small and dim IR targets using a fusion-based approach that incorporates a transformer attention module and an adaptive asymmetric fusion module. The overall structure of the Global Attention Network (GANet) is illustrated in Figure~\ref{GANet}, showcasing multiscale feature fusion. This network comprises distinct stages, namely feature extraction, transformer attention, adaptive asymmetric fusion, and up-sampling. Feature extraction is utilized to capture multiscale features from the input image, employing ResNet34 \cite{liang2022segmentation} pre-trained on ImageNet as the baseline, which includes Res1, Res2, and Res3 downsampling layers. To address the challenge of small infrared targets consisting of only a few pixels, the authors limit the increase in downsampling layers to prevent the loss of target information in deeper layers, thereby preserving detection accuracy for small targets. The feature maps obtained from various layers undergo processing in the transformer attention module (Trans1, Trans2, and Trans3). This module leverages global features to establish long-range dependencies for targets. An Adaptive Asymmetric Fusion (AAF) module integrates low-level and high-level features into the up-sampling stages, ensuring precise detection of small targets. The end-to-end network produces a binary map indicating predicted target locations, with dimensions identical to the input image.

Table  \ref{tab:deep}  showcases the evaluation of various deep learning methods across multiple datasets.  Given the diverse range of performance measures, datasets and hyperparameters under consideration, it is impractical to directly evaluate the performance of these aspects on a common platform. Nevertheless, it is usually evident that most of these techniques demonstrate significant efficacy in circumstances including both cluttered and homogeneous backgrounds. The majority of the techniques under consideration are founded on the re-engineering of downsampling methods in order to prevent the loss of small targets in deeper layers. A small subset of individuals have incorporated attention modules into their network architectures. The majority of the approaches exhibit exceptional performances. In addition to target detection, ISNet also offers contour information to a significant degree.

\begin{table*}[t]
\caption{Performance comparison of deep learning-based small and dim target detection methods.}
\label{tab:deep}
\resizebox{2\columnwidth}{!}{%
\begin{tabular}{@{}lllllll@{}}
\toprule
\multicolumn{1}{c}{Reference} &
  \multicolumn{1}{c}{Method} &
  \multicolumn{1}{c}{Type} &
  \multicolumn{1}{c}{Dataset} &
  \multicolumn{1}{c}{ } &
  Result &
   \\ \midrule
 \cite{mou2023yolo}  & 
  YOLO-FR &
  \begin{tabular}[c]{@{}l@{}}  Fully Supervised, Detection Type\end{tabular} &
  \cite{hui2019dataset} &
  Object Level &
  Precision (\%) &   97 \\
 &        &                                                                                 &            &              & Recall (\%)    & 95.4   \\
 &        &                                                                                 &            &              & mAP(50\%)      & 97.4   \\
 &        &                                                                                 &            &              & mAP(50-90\%)   & 82.6   \\ 
 \\ \midrule
\cite{dai2021asymmetric} & ACM    & \begin{tabular}[c]{@{}l@{}}  Fully Supervised\\ Segmentation Type\end{tabular}  & NUAA-SIRST & Pixel Level  & IOU            & 64.92  \\
 &        &                                                                                 &            & Object Level & Pd             & 90.87  \\
 &        &                                                                                 &            &              & Fa             & 12.76  \\  \\
 &        &                                                                                 & NUDT-SIRST & Pixel Level  & IOU            & 57.42  \\   \\
 &        &                                                                                 &            & Object Level & Pd             & 91.75  \\
 &        &                                                                                 &            &              & Fa             & 39.73  \\  \\
 &        &                                                                                 & IRSTD-1K   & Pixel Level  & IOU            & 57.42  \\  \\
 &        &                                                                                 &            & Object Level & Pd             & 91.75  \\
 &        &                                                                                 &            &              & Fa             & 39.73  \\  \\
 \\ \midrule
 \cite{dai2021attentional}& ALCnet & \begin{tabular}[c]{@{}l@{}}  Fully Supervised\\ Segmentation Type\end{tabular}  & NUAA-SIRST & Pixel Level  & IOU            & 67.91  \\
 &        &                                                                                 &            & Object Level & Pd             & 92.78  \\
 &        &                                                                                 &            &              & Fa             & 37.04  \\ \\
 &        &                                                                                 & NUDT-SIRST & Pixel Level  & IOU            & 61.78  \\ \\
 &        &                                                                                 &            & Object Level & Pd             & 91.32  \\
 &        &                                                                                 &            &              & Fa             & 36.36  \\ \\
 &        &                                                                                 & IRSTD-1K   & Pixel Level  & IOU            & 62.03  \\ \\
 &        &                                                                                 &            & Object Level & Pd             & 91.75  \\
 &        &                                                                                 &            &              & Fa             & 42.46  \\  \\
 \\\midrule
 \cite{li2022dense} & DNAnet & \begin{tabular}[c]{@{}l@{}} Fully Supervised\\ Segmentation Type\end{tabular}  & NUAA-SIRST & Pixel Level  & IOU            & 76.86  \\
 &        &                                                                                 &            & Object Level & Pd             & 96.96  \\
 &        &                                                                                 &            &              & Fa             & 22.5   \\ \\
 &        &                                                                                 & NUDT-SIRST & Pixel Level  & IOU            & 87.42  \\ \\
 &        &                                                                                 &            & Object Level & Pd             & 98.31  \\ 
 &        &                                                                                 &            &              & Fa             & 24.5   \\  \\
 &        &                                                                                 & IRSTD-1K   & Pixel Level  & IOU            & 62.73  \\ \\
 &        &                                                                                 &            & Object Level & Pd             & 93.27  \\
 &        &                                                                                 &            &              & Fa             & 21.81  \\  \\
 \\ \midrule

\end{tabular}
}
\end{table*}

 \begin{table*}[t]
 \resizebox{2\columnwidth}{!}{%
\begin{tabular}{@{}lllllll@{}}
\toprule
\multicolumn{1}{c}{Reference} &
  \multicolumn{1}{c}{Method} &
  \multicolumn{1}{c}{Type} &
  \multicolumn{1}{c}{Dataset} &
  \multicolumn{1}{c}{ } &
  Result &
   \\ \midrule
  
    \cite{wang2019miss}& MDvsFA & \begin{tabular}[c]{@{}l@{}} Fully Supervised\\ Segmentation Type\end{tabular}  & NUAA-SIRST & Pixel Level  & IOU            & 61.77  \\ 
 &        &                                                                                 &            & Object Level & Pd             & 92.40  \\
 &        &                                                                                 &            &              & Fa             & 64.90  \\  \\
 &        &                                                                                 & NUDT-SIRST & Pixel Level  & IOU            & 45.38  \\ \\
 &        &                                                                                 &            & Object Level & Pd             & 86.03  \\
 &        &                                                                                 &            &              & Fa             & 200.71 \\  \\
 &        &                                                                                 & IRSTD-1K   & Pixel Level  & IOU            & 35.40  \\ \\
 &        &                                                                                 &            & Object Level & Pd             & 85.86  \\
 &        &                                                                                 &            &              & Fa             & 99.22  \\  \\ \midrule
    \cite{zhang2022isnet} &
  ISNet &
  \begin{tabular}[c]{@{}l@{}} Fully Supervised\\ Segmentation Type\end{tabular} &
  NUAA-SIRST &
  Pixel Level &
  IOU &
  72.04 \\
 &        &                                                                                 &            & Object Level & Pd             & 94.68  \\
 &        &                                                                                 &            &              & Fa             & 42.46  \\ \\
 &        &                                                                                 & NUDT-SIRST & Pixel Level  & IOU            & 71.27  \\ \\
 &        &                                                                                 &            & Object Level & Pd             & 96.93  \\
 &        &                                                                                 &            &              & Fa             & 96.84  \\  \\
 &        &                                                                                 & IRSTD-1K   & Pixel Level  & IOU            & 60.61  \\ \\
 &        &                                                                                 &            & Object Level & Pd             & 94.28  \\
 &        &                                                                                 &            &              & Fa             & 61.28  \\  \\ \midrule
 \cite{peng2023dynamic}&
  DBR &
  \begin{tabular}[c]{@{}l@{}} Fully Supervised\\ Segmentation Type\end{tabular} &
  MFIRST &
  Object Level &
  Precision (\%) &
  63.80 \\
 &        &                                                                                 &            &              & Recall (\%)    & 72.24  \\
 &        &                                                                                 &            &              & F1             & 64.10  \\ \\
 &        &                                                                                 & SIRST      & Object Level & Precision (\%) & 80.70  \\
 &        &                                                                                 &            &              & Recall (\%)    & 74.09  \\
 &        &                                                                                 &            &              & F1             & 75.01  \\ \\ \midrule
\cite{wang2022mpanet} & MPAnet & \begin{tabular}[c]{@{}l@{}} Fully Supervised\\ Segmentation Type\end{tabular}  & SIRST      & Pixel Level  & IOU            & 74.5   \\
 &        &                                                                                 &            &              & nIOU           & 75.09  \\ \\
 &        &                                                                                 &            & Object Level & Pd             & 97.14  \\
 &        &                                                                                 &            &              & Fa             & 9.87   \\
 &        &                                                                                 &            &              & F1             & 0.8282 \\ \\ \midrule
 \cite{liu2021infrared}&
  Transformer based &
  \begin{tabular}[c]{@{}l@{}} Fully Supervised\\ Segmentation Type\end{tabular} &
  MFIRST &
  Pixel Level &
  F1 &
  64.59 \\
 &        &                                                                                 &            & Object Level & Pd             & 90.08  \\
 &        &                                                                                 &            &              & F1             & 92.59  \\ \\
 &        &                                                                                 & SIRST      & Pixel Level  & F1             & 83.16  \\ \\
 &        &                                                                                 &            & Object Level & Pd             & 100    \\
 &        &                                                                                 &            &              & F1             & 98.62  \\ \\ \midrule

\end{tabular}
}
\end{table*}

 \begin{table*}[t]
 \resizebox{2\columnwidth}{!}{%
\begin{tabular}{@{}lllllll@{}}
\toprule
\multicolumn{1}{c}{Reference} &
  \multicolumn{1}{c}{Method} &
  \multicolumn{1}{c}{Type} &
  \multicolumn{1}{c}{Dataset} &
  \multicolumn{1}{c}{ } &
  Result &
   \\ \midrule

\cite{zhang2024global} & Ganet  & \begin{tabular}[c]{@{}l@{}} Fully Supervised
Segmentation Type\end{tabular} & SIRST & Pixel Level  & IOU            & 46.29  \\ \\
 &        &                                                                                 &            & Object level & AUC             & .9447  \\
 &        &                                                                                 &            &              & F1             & 68.58  \\  \\ \midrule   
\cite{ying2023mapping} & LESPS  & \begin{tabular}[c]{@{}l@{}} Weakly Supervised\end{tabular} & NUAA-SIRST & Pixel Level  & IOU            & 61.13  \\ \\
 &        &                                                                                 &            & Object level & Pd             & 93.16  \\
 &        &                                                                                 &            &              & Fa             & 11.87  \\ \\
 &        &                                                                                 & NUDT-SIRST & Pixel Level  & IOU            & 58.37  \\ \\
 &        &                                                                                 &            & Object level & Pd             & 93.76  \\
 &        &                                                                                 &            &              & Fa             & 28.01  \\ \\
 &        &                                                                                 & IRSTD-1K   & Pixel Level  & IOU            & 49.05  \\  \\
 &        &                                                                                 &            & Object level & Pd             & 87.54  \\
 &        &                                                                                 &            &              & Fa             & 15.07  \\ \\ \bottomrule
\end{tabular}
}
\end{table*}

\subsubsection{Weakly Supervised Learning Based Approaches}

Training CNNs for the detection of IR small targets using fully supervised learning has gained significant attention recently. However, these methods are resource-intensive because they require a substantial amount of manual effort to annotate each pixel.

\par As presented in \cite{ying2023mapping}, authors have introduced an innovative solution to address this challenge by using point-level supervision. During the training process, CNNs are guided by point labels. Interestingly, CNNs initially learn to segment a cluster of pixels near the targets. Gradually, they refine their predictions to closely match the ground truth point labels. The approach described in \cite{ying2023mapping} is influenced by the notion of \textit{mapping degeneration} and is implemented through a framework called Label Evolution with Single Point Supervision (LESPS). One of the key motivations behind their research arises from an intriguing observation made while training SIRST detection networks. When single-point labels are employed for guidance, CNNs tend to initially segment a group of pixels near the targets with low confidence. However, over time, the CNNs progressively enhance their performance and exhibit higher confidence in predicting accurate ground truth point labels. 
The process of \textit{mapping degeneration} is shaped by the unique imaging attributes of IR systems, as shown in studies \cite{li2022dense, zhang2022isnet}, as well as by the local contrast properties of small IR targets. Additionally, the inherent learning progression of CNNs, as discussed in \cite{ulyanov2018deep}, plays a role in causing this degeneration. The first two factors lead to the enlargement of segmented areas beyond the defined point labels, while the third factor contributes to this degeneration process. They introduced the LESPS framework for weakly supervised SIRST detection. LESPS leverages intermediate predictions generated by the neural network during training to iteratively update the current labels. These updated labels serve as supervision until the next label update. By employing iterative label updates and network training techniques, the neural network can approximate these updated pseudo mask labels, enabling end-to-end pixel-level SIRST detection.

\section{Popular Datasets } \label{sec:four}
\begin{figure}[t]
	\centering
	\includegraphics[scale=0.30]{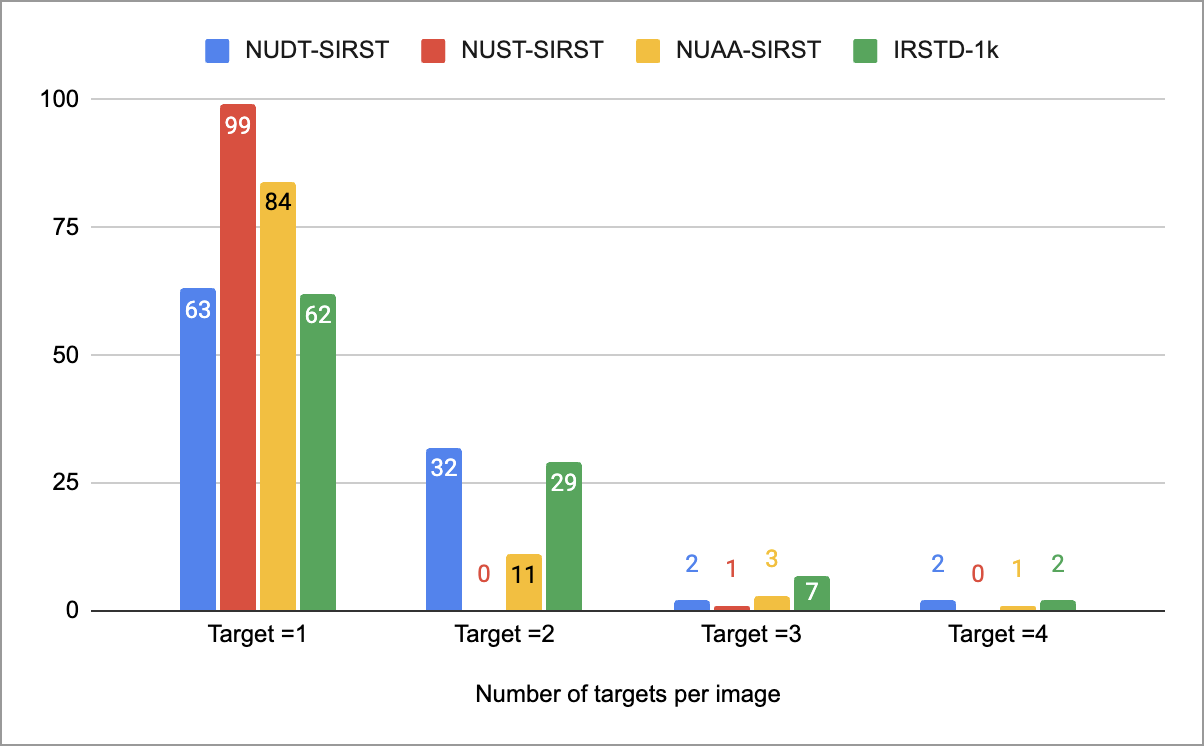}

	\caption{Distribution of targets in various datasets.}
	\label{dgr1}
\end{figure}

\begin{figure}[t]
	\centering
	\includegraphics[scale=0.30]{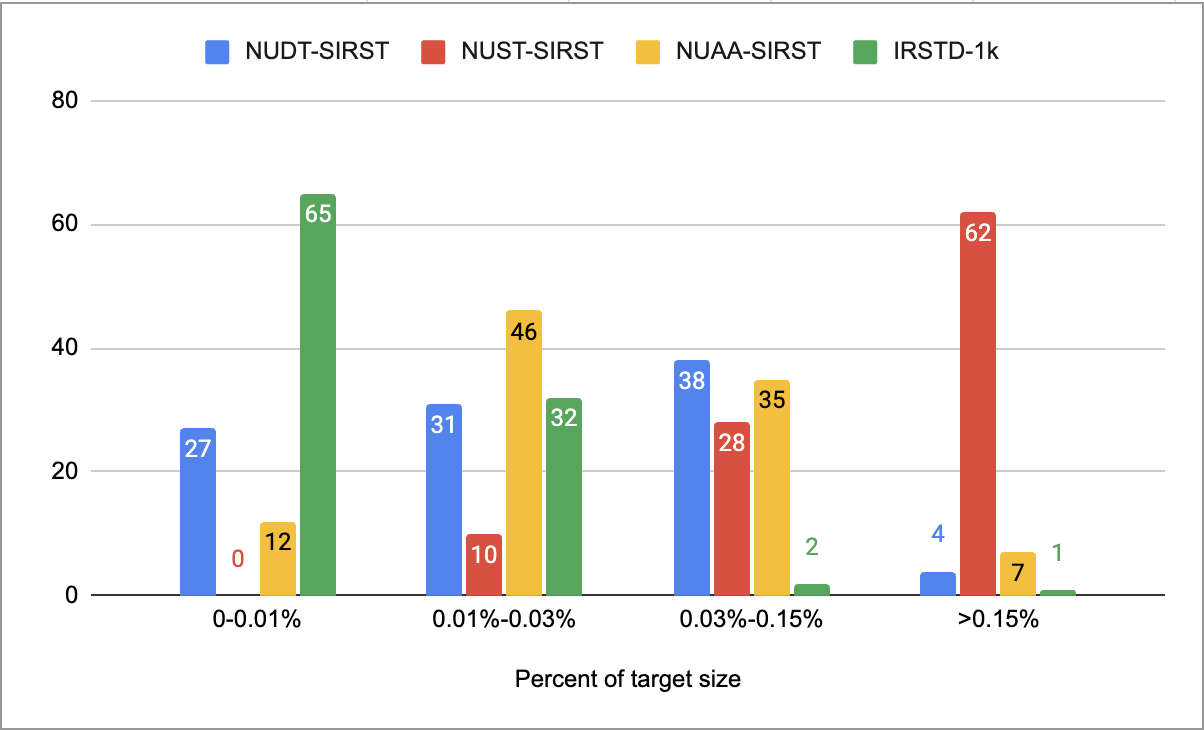}
	\caption{Target sizes in various datasets.}
	\label{dgr2}
\end{figure}

\begin{figure}[t]
	\centering
	\includegraphics[scale=0.30]{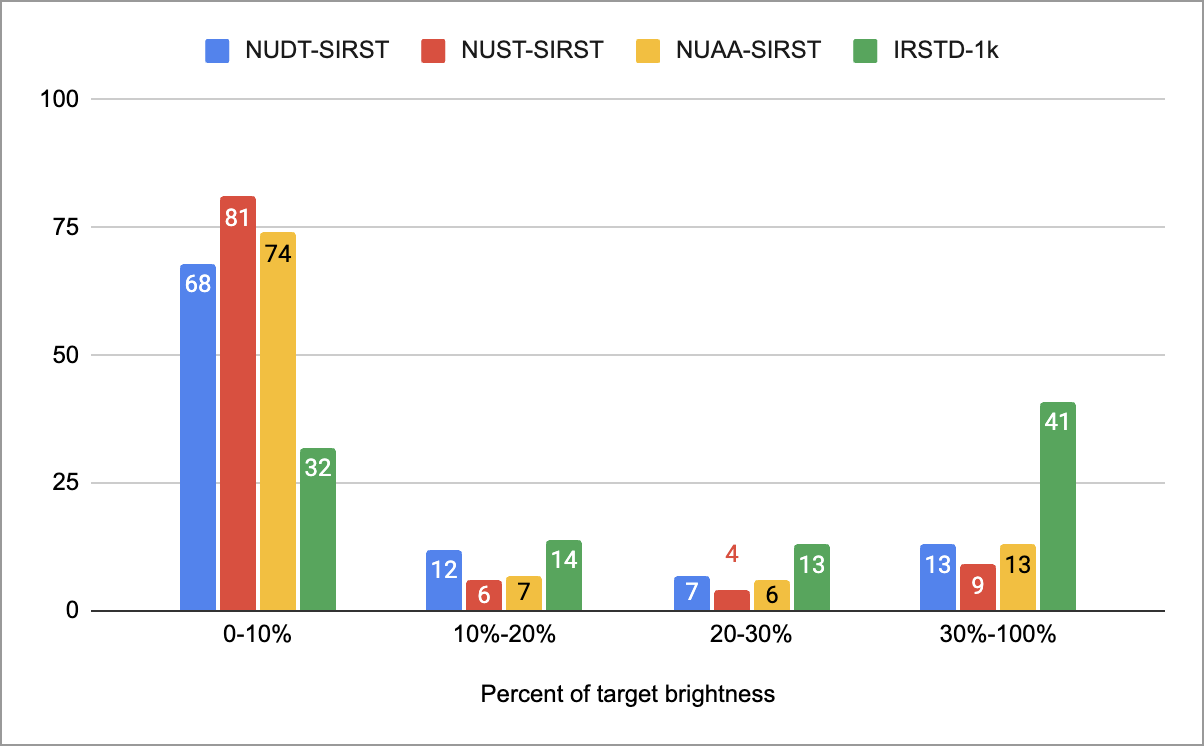}
	\caption{Target brightness in various datasets.}
	\label{dgr3}
\end{figure}

The lack of large-scale datasets presents a significant obstacle to the adoption of deep learning methods for detecting small and dim targets in the IR domain. Although there has been some progress in recent years in dataset generation (see Table~\ref{tab:dtst} and Figures~\ref{dgr1}, \ref{dgr2}, and \ref{dgr3}) related to this area, the subsequent section provides an overview of the datasets currently available in this area.

\subsection{SIRST Dataset}

The SIRST  dataset plays a pivotal role in the realm of small and dim target detection within IR imagery. It functions as a benchmark dataset, uniquely tailored to the evaluation and advancement of algorithms crafted specifically for the purpose of detection, tracking, and classification of small targets within IR environments.

The SIRST dataset offers a diverse array of IR image sequences procured from a variety of platforms, encompassing both aerial and ground-based sensors. These sequences span diverse scenarios, including urban environments, rural landscapes, and maritime settings. The dataset covers an extensive spectrum of target dimensions, signal-to-noise ratios, and background intricacies, effectively representing authentic and intricate conditions for the task of target detection. The SIRST framework includes the following several prominent datasets that have garnered considerable acclaim.
\subsubsection{NUAA-SIRST Dataset}
This dataset \cite{dai2021asymmetric} includes real IR images featuring a wide array of backgrounds, including scenes like clouds, urban settings, and bodies of water. It comprises a total of 427 images, which have been meticulously annotated. Within this dataset, one can find a variety of targets, ranging from small and dim point targets to extended targets.
\subsubsection{NUST-SIRST Dataset}
This dataset \cite{wang2019miss} holds a prominent position in the IR imaging domain and classifies targets into two main categories: point and spot types. It consists of an extensive collection of 10,000 images, encompassing diverse backgrounds, such as clouds, cityscapes, rivers, and roadways. The dataset has been annotated manually with a broad classification. It's worth noting that this dataset is artificially generated using synthetic techniques.
\subsubsection{CQU-SIRST Dataset}
This dataset \cite{gao2013infrared} includes a collection of 1676 synthetic images. It incorporates diverse background scenarios, encompassing settings like clouds, urban environments, and maritime scenes. The dataset also includes corresponding ground truth data. The main focus of this dataset is on point targets.
\subsubsection{NUDT-SIRST Dataset}
The dataset \cite{li2022dense} comprises a sum of 1327 synthetically generated images. It covers a wide range of background scenarios, including settings like clouds, urban environments, and maritime scenes. Ground truth data is also included along with the dataset. It contains a mixture of point and extended targets.

\subsection{IRSTD Dataset}
The IRSTD-1K \cite{zhang2022isnet} dataset is a recent dataset where the collection of 1,000 IR images was obtained in real-world scenarios using an IR camera. These images have dimensions of 512×512 pixels. To ensure precise annotations, the targets within these images have been meticulously labeled at the pixel level. Within the IRSTD-1K dataset, a variety of small targets are included, encompassing categories such as drones, creatures, vessels and vehicles. These targets have been captured at different locations, and the images were acquired from considerable imaging distances. The dataset spans a diverse array of environments, including scenes involving bodies of water, natural landscapes, urban settings, and atmospheric conditions.

The backgrounds in the IRSTD-1K dataset exhibit notable clutter and noise, further enhancing the complexity of detecting and recognizing targets in IR imagery. Consequently, this dataset is well-suited for comprehensive assessments and method benchmarking, specifically in the domain of IR small target detection (IRSTD). The IRSTD-1K benchmark provides a valuable resource for researchers, enabling them to evaluate and compare the performance of their algorithms and models for the accurate detection and classification of small targets in IR images.

\subsection{Customized Dataset}
In addition to the aforementioned datasets, numerous researchers \cite{liu2017image} have created their own unique datasets \cite{10249124, kumar2021detection} specifically designed for the purpose of evaluating their algorithms. Naraniya et al. \cite{naraniya2021scene} presented a methodology for creating a specialized dataset by combining artificially generated backgrounds with synthetic targets. In this methodology, the motion of the target is modeled in the NED coordinate system. Additionally, Gaussian blurred point targets are superimposed onto these backgrounds.

\begin{table*}[t]
\caption{Details of various available datasets.}
\label{tab:dtst}
\resizebox{2\columnwidth}{!}{%
\begin{tabular}{@{}ccccccc@{}}
\toprule
\textbf{Dataset} &
  \textbf{Image Type} &
  \textbf{Background Scene} &
  \textbf{\begin{tabular}[c]{@{}c@{}}Number of\\ Images\end{tabular}} &
  \textbf{Label Type} &
  \textbf{Target Type} &
  \textbf{Availability} \\ \midrule
\textbf{NUAA-SIRST \cite{dai2021asymmetric}} &
  Real &
  Cloud/City/Sea &
  427 &
  \begin{tabular}[c]{@{}c@{}}Manual Coarse\\ Label\end{tabular} &
  Point/Spot/Extended &
  Public \\ \\ \midrule
\textbf{NUST-SIRST \cite{wang2019miss}} &
  Synthetic &
  Cloud/City/River/Road &
  10,000 &
  \begin{tabular}[c]{@{}c@{}}Manual Coarse\\ Label\end{tabular} &
  Point Spot &
  Public \\ \\ \midrule
\textbf{CQU-SIRST \cite{gao2013infrared}} &
  Synthetic &
  Cloud/City/Sea &
  1676 &
  Ground Truth &
  Point Spot &
  Private \\ \\ \midrule
\textbf{NUDT-SIRST \cite{li2022dense}} &
  Synthetic &
  Cloud/City/Sea/Highlight/Field &
  1327 &
  Ground Truth &
  Point/Spot/Extended &
  Public \\ \\ \midrule
\textbf{IRSTD \cite{zhang2022isnet}} &
  Real &
  Sea/River/Field/Mountain/City/Cloud &
  1000 &
  Ground Truth &
  Point/Spot/Extended &
  Public \\ \\ \bottomrule
\end{tabular}
}
\end{table*}

\section{Performance Evaluation} \label{sec:five}
Commonly used metrics to evaluate the performance of methods for small and dim target detection in IR Imagery are discussed below.
\subsection{Performance Measure at Pixel level}
Intersection over Union (IoU) is a metric commonly used in computer vision and object detection tasks. It measures the overlap between two bounding boxes or regions of interest (ROIs). IoU is a metric used to evaluate the accuracy of an object detection algorithm. IoU is calculated by finding the ratio of the area of overlap between the predicted bounding box and the ground truth bounding box to the area of their union as follows:
\begin{equation}
            {IoU= \frac{A_i}{A_u}}
\end{equation}            
            
where $A_i$ represents the area of the intersection region and $A_u$ represents the area of the union region. 

The IoU value ranges from 0 to 1, where 0 indicates no overlap between the predicted and ground truth bounding boxes, and 1 indicates a perfect match. Normalized Intersection over Union $(nIoU)$ is a metric that is typically expressed as the Intersection over Union (IoU) divided by the sum of the areas of the predicted bounding box and the ground truth bounding box minus IoU. It is a normalized version of the Intersection over Union $(IoU)$ metric.
\begin{equation}
nIoU=\frac{1}{N}\sum_{i=1}^{N}\left ( \frac{TP[i]}{\left (T[i]+P[i] - TP[i]  \right )} \right )
\end{equation}
In the above equation, $N$ represents the total number of samples. $TP[i]$ represents the count of true positive pixels, while $T[i]$ and $P[i]$ represent the count of ground truth and predicted positive pixels, respectively.

\subsection{Performance Measure at Object level}

At the object level, accurate detection is considered when both of the following conditions are met simultaneously: the resulting output shows a degree of pixel overlap with the ground truth, and the gap in pixels between the centers of the detection result and the ground truth is below a specified threshold. The Probability of Detection $(P_d)$ is a statistical measure that quantifies the likelihood of correctly identifying the presence of a target or signal in a given system or scenario.

\par $P_d$, also known as the Probability of Detection, is a metric that represents the ratio of correctly predicted targets $(N_{\text{pred}})$ to all targets $(N_{\text{all}})$.

\begin{equation}
P_{d}=\frac{N_{pred}}{N_{all}}
\end{equation}

False-Alarm Rate $(F_a)$ is the ratio of the number of falsely predicted target pixels $(N_{false})$ and the total number of pixels in the image $(N_{all})$.

\begin{equation}
F_{a}=\frac{N_{false}}{N_{all}}
\end{equation}

\begin{table}[t]
\caption{Confusion matrix.}
\label{tab:cnf}
\begin{tabular}{@{}lll@{}}
\toprule
\textbf{} & \textbf{True} & \textbf{False}         \\ \midrule
Positive  & TP            & FP                     \\
Negative  & TN            & \multicolumn{1}{c}{FN} \\ \bottomrule
\end{tabular}
\end{table}

Other evaluation measures employed for assessing model performance include precision, recall, and mean average precision (mAP). The calculation of both precision and recall relies on the utilization of the confusion matrix, as depicted in Table \ref{tab:cnf}. As shown in Equation \ref{eq:prc}, Precision is calculated as the ratio of true positives to the sum of true positives and false positives. Precision is the number of true positive predictions divided by the total number of positive predictions made by the model. In the context of object detection, precision is the ratio of correctly detected objects to the total number of objects predicted by the model. As shown in Equation~\ref{eq:rcl}, Recall is calculated as the ratio of true positives to the sum of true positives and false negatives. Recall is the number of true positive predictions divided by the total number of actual positive instances. In object detection, recall is the ratio of correctly detected objects to the total number of ground truth objects. In the context of target detection, there is often a trade-off between precision and recall. Increasing one metric may result in a decrease in the other. Therefore, it's common to use a combined metric like \(F1\) score, which considers both precision and recall. \(F1\) score is defined in Equation \ref{eq:f1}. Average Precision (AP) is calculated by taking the precision-recall curve of a model's predictions and computing the area under that curve. It summarizes the precision-recall trade-off for different confidence thresholds. Mean Average Precision (mAP) is the mean of the Average Precisions across multiple classes or categories. It is often used in scenarios where there are multiple classes, and we want to assess the overall performance of the model. Each class has its own precision-recall curve and AP, and mAP provides an aggregated performance measure, as represented by Equation \ref{eq:apl}.

\begin{equation}
 Precision = \frac{TP}{TP + FP}
 \label{eq:prc}
 \end{equation}
 
 \begin{equation}
 Recall = \frac{TP}{TP + FN}
 \label{eq:rcl}
 \end{equation}

 \begin{equation}
F_{1}=\frac{2*Precision*Recall}{Precision+Recall}
 \label{eq:f1}
 \end{equation}

\begin{equation}
mAP = \frac{1}{N} \sum_{i=1}^{N} AP_i
 \label{eq:apl}
 \end{equation}

Where $N$ is the number of classes, and $AP_i$ is the Average Precision for class $i$. In the context of object detection, a higher mAP indicates better performance, meaning that the model is good at both correctly identifying objects (precision) and not missing any objects (recall) across multiple classes.

\section{Discussion and Potential Future Directions} \label{sec:six}

\par Detecting small and dim targets using MIRST methods requires multiple frames, making them less suitable for real-time implementation. Approaches based on low-rank representation have demonstrated adaptability to low signal-to-clutter ratio IR images. However, these algorithms still encounter challenges in accurately detecting small and variably shaped targets in complex backgrounds, resulting in a high False Alarm Rate (FAR).

HVS-based approaches involve assessing the discrepancy between targets and backgrounds through discontinuity measurement. In cluttered scenarios with low signal-to-noise ratio (SNR), these algorithms exhibit inadequate performance. Traditional image processing relies on manually designed features, but the significant variations in real scenes (such as target size, shape, signal-to-clutter ratio (SCR), and clutter backdrop) make using handcrafted features and fixed hyperparameters challenging in effectively addressing these differences. Achieving a generalized implementation without scenario-specific hyperparameter tuning is highly challenging. The complexity of real scenarios, marked by dynamic changes in target size, shape, and cluttered background, poses challenges in utilizing handcrafted features and fixed hyperparameters to address these fluctuations. Additionally, incorporating handcrafted features and optimizing hyperparameters require specialized expertise and substantial engineering efforts. Most conventional image processing-based algorithms perform effectively on images with uniform backgrounds and distinct contrast targets. However, their performance is suboptimal in the presence of significantly diverse cluttered backgrounds.

Conventional small and dim target detection methods exhibit a susceptibility to generate instances of both missed detections and false detections in scenarios where the signal-to-clutter ratio (SCR) is low. Additionally, these methods tend to produce false alarm detections in situations characterized by high local contrast. In contrast to conventional image processing approaches, deep learning-based methodologies have the capability to acquire small and dim IR target characteristics through a data-centric approach. While deep learning-based methods have recently demonstrated state-of-the-art performance, it is important to note that the majority of these methods have only fine-tuned networks that were originally built for generic extended objects. The application of these approaches for small target detection in IR  can result in the loss of small targets in deeper layers due to the significant size difference between IR small targets and generic objects.

Deep Learning based approaches typically exhibit superior performance compared to traditional methods. However, in general, these may encounter limitations in accurately predicting target shapes. The robust training of deep learning-based techniques necessitates a large-scale dataset due to its data-centric nature. The effective training of deep learning techniques requires a large dataset due to its emphasis on data-driven approaches. Segmentation-type approaches necessitate a dataset with pixel-level annotations, which entails a substantial amount of engineering work. In contrast, detection-type approaches and weakly supervised approaches require less engineering effort for dataset generation.

As technology continues to advance, several potential future directions can be explored to enhance the accuracy and efficiency of detecting small and dim targets in IR imagery. One promising avenue for improvement lies in the development of advanced machine learning algorithms, particularly deep learning models, tailored specifically for small target detection. Convolutional Neural Networks (CNNs) have shown success in various computer vision tasks, and their application to small target detection in IR imagery can be further refined. Researchers may explore novel network architectures, optimization techniques, and training strategies to boost the model's ability to discern subtle features indicative of small and dim targets. In addition to traditional supervised learning approaches, there is potential for leveraging unsupervised and semi-supervised learning techniques. Anomaly detection methods, which can identify deviations from normal patterns in the absence of labeled training data, could be explored to enhance small target detection in situations where annotated datasets are limited. Incorporating transfer learning from other domains or modalities may also prove beneficial in training models with limited IR imagery datasets.

The integration of multispectral and hyperspectral data is another promising direction \cite{gallagher2023assessing}. Combining information from different wavelengths beyond the infrared spectrum can provide a more comprehensive view of the scene, aiding in the discrimination of small and dim targets from background noise. Fusion techniques that intelligently merge data from multiple sensors or platforms can be explored to exploit the complementary strengths of different spectral bands. Advancements in sensor technology, including higher spatial and temporal resolutions, could significantly impact small target detection. The development of sensors with improved sensitivity and dynamic range, as well as the incorporation of emerging technologies such as quantum sensors, may enhance the overall quality of IR imagery, making it easier to identify small and dim targets with greater accuracy. Furthermore, the exploration of real-time processing capabilities is essential for applications where timely detection is critical. Implementing parallel processing techniques, leveraging Graphics Processing Units (GPUs) or dedicated hardware accelerators, can expedite the analysis of large volumes of IR imagery in real time, ensuring swift and accurate detection of small and dim targets.

\section{Conclusion} \label{sec:seven}
This review encompasses a wide variety of approaches developed and refined over the years for small and dim target detection in IR imagery. The authors aggregated findings and conducted a comparative analysis of the majority of approaches in tabular form. It can be inferred that most deep learning-based methods for detecting small and dim targets exhibit notable performance, demonstrating a discernible improvement compared to conventional image processing-based approaches, especially in scenarios with cluttered backgrounds. Traditional image processing methods, specifically MTHM and CM, exhibit commendable performance and have a minimal computational footprint. However, they lack the capability to manage cluttered backgrounds effectively. In contrast, deep learning-based techniques show exceptional performance in both cluttered and uniform backgrounds.

\section{Acknowledgement} \label{sec:eight}
Throughout the duration of this undertaking, the authors wish to extend their heartfelt gratitude to Ms.Neeta Kandpal, Scientist 'G', IRDE, Dehradun, and Dr. Ajay Kumar, Outstanding Scientist and Director, IRDE, Dehradun, for their invaluable support and encouragement. The realization of this groundbreaking endeavor would not have been possible in the absence of his guidance.

\bibliographystyle{unsrtnat}
\bibliography{cas-refs}

\newpage

\bio{nikhil}
Nikhil Kumar received his M.Tech. degree from the Department of Electrical Engineering, IIT Kanpur. He is currently pursuing a Ph.D. from the Department of Computer Science, IIT Roorkee. As a scientist at DRDO, he has extensive experience in developing electro-optical systems for the Indian Armed Forces. His area of research is IR Signal Processing, Image processing, Computer Vision, Deep Learning, and Artificial Intelligence. 
\endbio

\bio{psingh}
Pravendra Singh received his Ph.D. degree from IIT Kanpur. He is currently an Assistant Professor in the CSE department at IIT Roorkee, India. His research interests include deep learning, machine learning, computer vision, and artificial intelligence. He has published papers at internationally reputable conferences and journals, including IEEE TPAMI, IEEE TIV, IJCV, CVPR, ECCV, NeurIPS, AAAI, IJCAI, Pattern Recognition, Neural Networks, Knowledge-Based Systems, Neurocomputing, and others.
\endbio

\end{document}